%% file: iclr2027_conference.tex
\documentclass{article} % For LaTeX2e
\usepackage{iclr2027_conference,times}

\input{math_commands.tex}

\usepackage{hyperref}
\usepackage{url}

\usepackage{wrapfig}
\usepackage{booktabs}
\usepackage{graphicx}
\usepackage{subcaption}
\usepackage{multirow}
\usepackage{float}
\usepackage{capt-of}

\usepackage[table]{xcolor}

\definecolor{ricoblue}{HTML}{E3F0FF}

\newcommand{\best}[1]{\textbf{#1}}
\newcommand{\second}[1]{\underline{\textit{#1}}}

\def\methodname{RiCo}

\title{\methodname: Neural Simulation of Rigid-Body Interactions via Local Contact Reasoning}

\author{Ruixiang Ouyang$^{1,2}$, Guanren Qiao$^1$, Fansen Meng$^4$, Yueci Deng$^{1,2}$, Ruixing Jin$^1$, 
\\ \textbf{Kui Jia}$^{1,2}$, \textbf{Guiliang Liu}$^{1,3}$\thanks{Corresponding author: Guiliang Liu, liuguiliang@cuhk.edu.cn} \\
$^{1}$The Chinese University of Hong Kong, Shenzhen\\
$^{2}$DexForce Co., Ltd.\\
$^{3}$Shenzhen Loop Area Institute \\
$^{4}$South China University of Technology \\
}

\iclrfinalcopy % Uncomment for camera-ready version, but NOT for submission.
\begin{document}

\maketitle
\vspace{-0.1in}
\begin{abstract}
Accurate simulation of rigid-body interactions is essential for predictive physical world models.
Despite recent progress in modeling object dynamics, capturing how local contacts between surfaces shape object motion remains challenging.
While end-to-end world models predict interactions across entire scenes or objects, in practice, rigid-body contact is inherently local, and only nearby surfaces can directly exchange contact forces.
Motivated by this observation, 
we introduce Rigid-body Contact Reasoning (RiCo), which represents interactions between objects through sparse neighborhoods of contact surface points.
\methodname\ combines each point's state with the relative geometry, motion, and physical properties of nearby surfaces, then reasons across the object's points to determine how these local contacts jointly affect its motion.
By confining cross-object reasoning to nearby surfaces while propagating contact information within each rigid body, \methodname\ retains fine-grained interaction details without the cost of modeling every pair of scene points. Such properties enable \methodname\ a higher accuracy and contact fidelity.
Experiments on MOVi-benchmark demonstrate that \methodname\ reduces 100-frame position and orientation errors by 31--35\% and approximately 38\%, respectively, compared with baselines. Moreover, \methodname\ achieves high contact fidelity, with ground-truth-relative penetration-time and mean-depth differences of 11.0\% and 2.22 mm, respectively. 
\methodname\ further generalizes zero-shot from small-scale training scenarios to scenes containing 270 objects. Our real-world multi-ball collision experiments further provide preliminary evidence of sim-to-real transfer.
\end{abstract}
\vspace{-0.1in}
\section{Introduction}
\vspace{-0.1in}
High-fidelity modeling of the physical world has become a central milestone in embodied AI, enabling agents to reason about how environments evolve and respond to their interactions~\citep{agarwal2025cosmos}.
Recent advances in world models have learned predictive representations from large-scale video observations~\citep{Bruce2024Genie,ye2026world,li2026causal}. 
Beyond these visual patterns, however, the physical world exhibits complex dynamics involving spatial motion, object geometry, and contact interactions~\citep{hafner2023mastering}, which remain difficult to capture with video generation backbones.
To advance the modeling of dynamic objects, we study neural simulation of rigid-body motions, including translation, rotation, and interactions among multiple objects in 3D space~\citep {chang2016compositional,Wei2025integrating}. 
As a fundamental abstraction of physical object motion, rigid-body dynamics exclude continuous deformation of objects, but introduce a significant challenge in modelling their interactions under different geometric form and motion conditions,
because collisions introduce abrupt changes in linear and angular velocities, making contact dynamics non-smooth and challenging to predict accurately~\citep{pfaff2020learning,Wei2025integrating}.

To address these challenges, prior works proposed learning neural approximations of motion dynamics from observed trajectories~\citep{hafner2019learning,brandstetter2022message,Wu2023SlotFormer}.
Despite their success across various domains, modeling rigid-body dynamics remains challenging because collisions may induce abrupt changes in objects' velocities and subsequent motion trajectories.
To better capture these discontinuities, recent studies incorporate geometric priors to explicitly model contact interactions. 
For instance, mesh-based neural simulators exploit mesh connectivity and surface geometry to propagate information about physical interactions~\citep{pfaff2020learning,allen2022learning}.
HOPNet~\citep{Wei2025integrating} further incorporates higher-order topological structures to improve the modeling of rigid-body contact dynamics.
However, processing mesh connectivity and constructing high-order geometric relationships can be computationally expensive and time-consuming.
Therefore, RigidFormer~\citep{dou2026rigidformer} represents individual objects using point sets and models their interactions through object-level attention, {enabling rigid-body simulation without explicit mesh connectivity}.
While point clouds can efficiently capture surface geometry, they do not inherently encode which surface points may come into contact during object interactions.

\begin{wrapfigure}{r}{0.6\columnwidth}
    \centering
    \vspace{-0.1in}

    \begin{subfigure}[t]{0.32\linewidth}
        \centering
        \includegraphics[width=\linewidth]{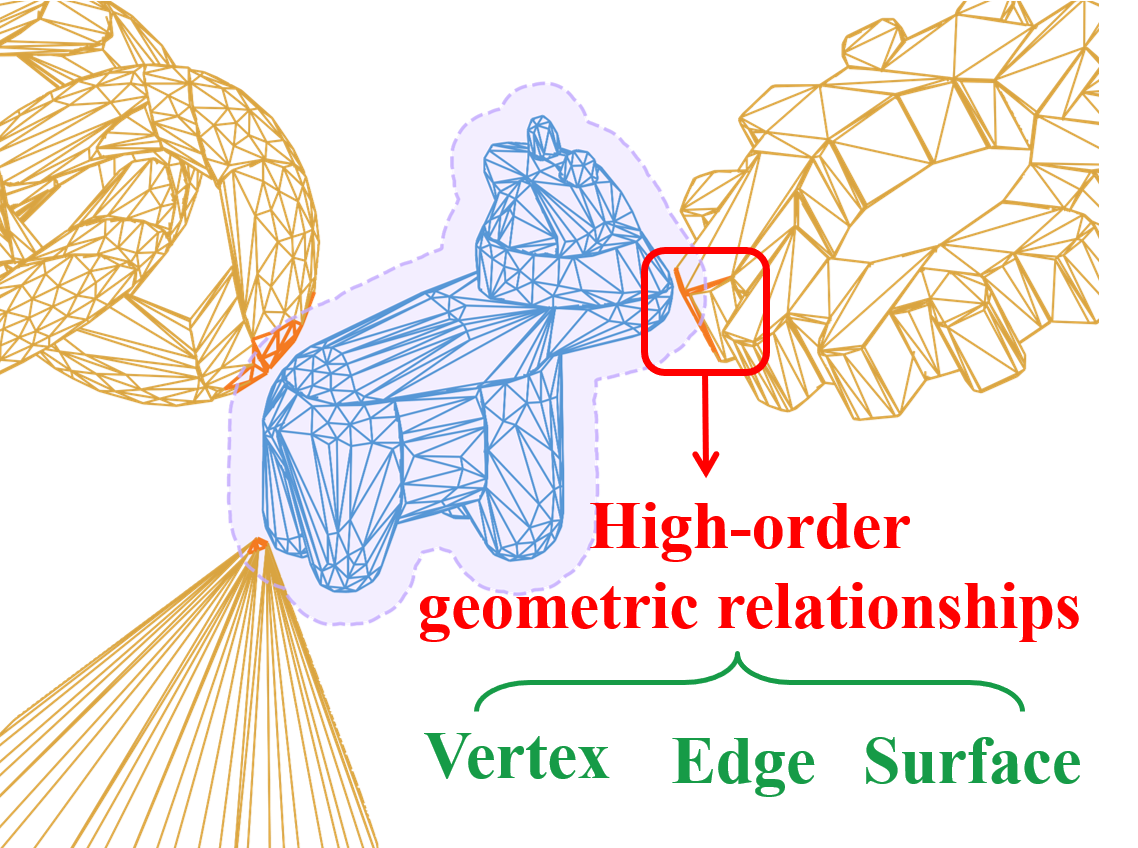}
        \caption{Mesh-based}
        \label{fig:motivation_a}
    \end{subfigure}
    \hfill
    \begin{subfigure}[t]{0.32\linewidth}
        \centering
        \includegraphics[width=\linewidth]{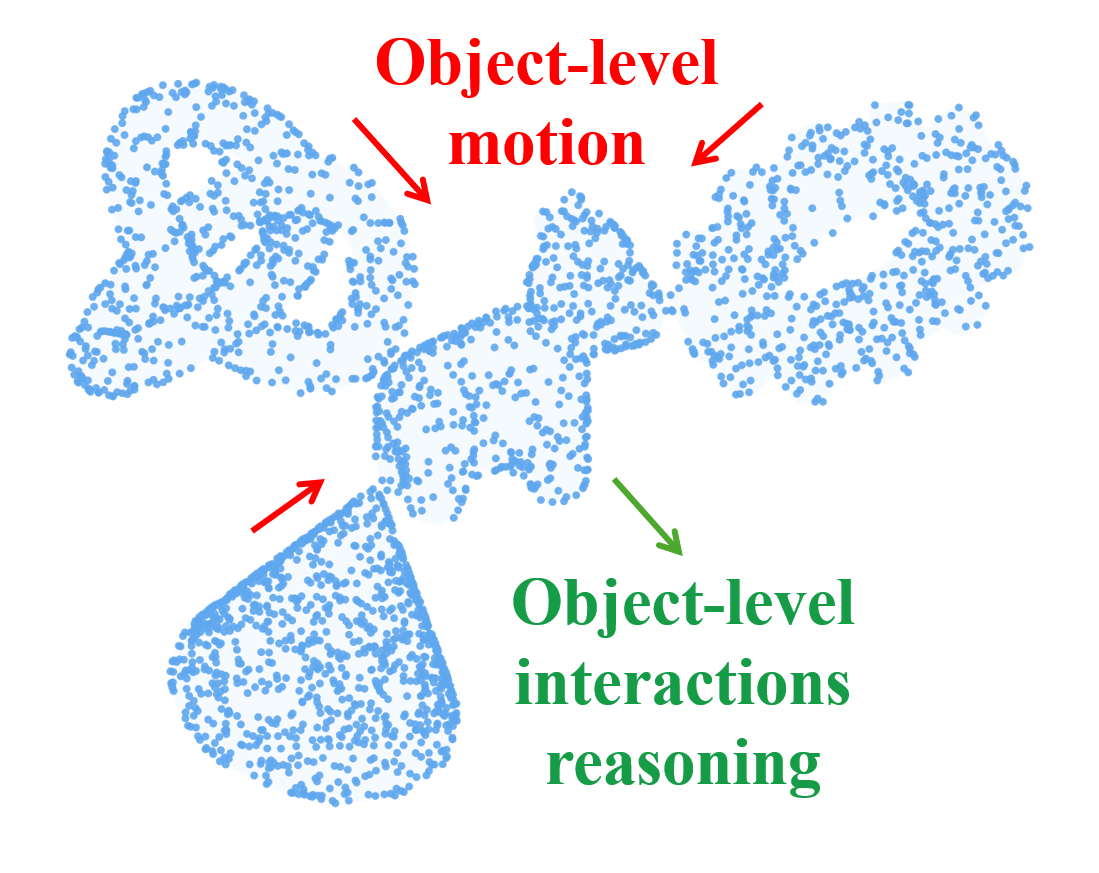}
        \caption{RigidFormer}
        \label{fig:motivation_b}
    \end{subfigure}
    \hfill
    \begin{subfigure}[t]{0.32\linewidth}
        \centering
        \includegraphics[width=\linewidth]{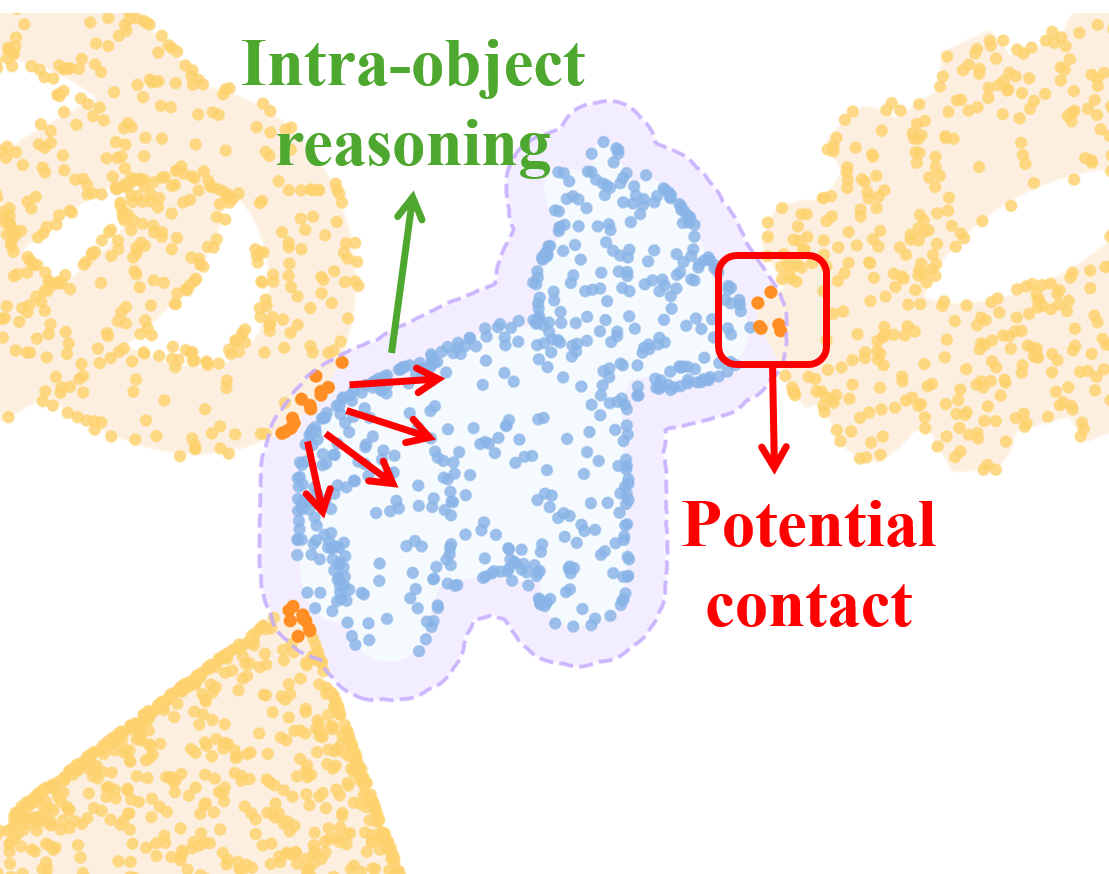}
        \caption{Ours}
        \label{fig:motivation_c}
    \end{subfigure}
    \caption{
    {Comparing rigid body contact modeling.}
    }
    \label{fig:motivation}
    \vspace{-0.1in}
\end{wrapfigure}

Striving for efficient and fine-grained contact modeling, we propose \methodname\, which identifies and models local contact relationships directly from point-cloud representations, without relying on mesh connectivity. Figure~\ref{fig:motivation} shows an illustrative example.
This approach reflects the local nature of rigid-body interactions: only nearby surfaces can directly exchange contact forces~\citep{sanchez2020learning}.
Accordingly, instead of modeling the dense interactions among all objects,
\methodname\ identifies only potential contact points and models their local dynamics across interacting objects.  
Because the combined effects of these contacts can govern rigid-body motion, we reason over contact information by propagating the effects of local contacts across rigid body.
The intra-object contact reasoning provides a principled link between fine-grained interactions and whole-object motion. This design not only supports generalization across diverse shapes but also scales efficiently as the number of surface points grows.

To implement \methodname, we build a neural simulator that decomposes object-level collisions into sparse local contacts and intra-object contact reasoning.
For each surface point, \methodname\ constructs a compact contact neighborhood containing nearby points from other objects and the environment.
It then combines the source point's state with the relative geometry, motion, and physical properties of these contact candidates to form contact-conditioned point features.
Based on such features, a point transformer attends each surface point to the others on the target object, so the model can reason where contacts occur, how the contacting surfaces move, and how forces at different locations affect translation and rotation.
Based on these contact-conditioned features, an anchor-based decoder predicts motion updates at a small set of surface points, from which a single rigid transformation is recovered and applied to the entire object.
This design integrates fine-grained geometric information into contact-dynamics reasoning, thereby generating physically plausible rigid-body motion.

We conduct extensive experiments on the MOVi benchmark~\citep{greff2022kubric} to evaluate \methodname\ in terms of prediction accuracy, generalization, {contact fidelity}, scalability, and real-world dynamics.
Across diverse object configurations and long-horizon rollouts, \methodname\ reduces position and orientation errors by up to 35\% and 38\%, respectively.
Beyond conventional trajectory metrics, we assess contact fidelity by comparing the predicted and ground-truth trajectories’ aggregate penetration-time ratios and mean penetration depths, thereby accounting for geometric overlap already present in the ground-truth trajectories.
Under this metric, \methodname\ reduces relative contact violations to 11.0\% of object-time steps, with an average relative penetration depth of 2.22\,mm.
Moreover, \methodname\ 
generalizes effectively across unseen distributions of 270 objects. 
We further investigate zero-shot sim-to-real transfer through real-world multi-ball billiard collision experiments, providing preliminary evidence that the learned contact dynamics can generalize to real-world multi-object interactions without additional training.
These results demonstrate that local contact reasoning enables accurate, contact-fidelity, and scalable neural simulation of complex rigid-body dynamics.
\vspace{-0.2in}
\section{Related work}
\label{sec:Related work}
\vspace{-0.15in}
\textbf{Learned Models of Physical Dynamics.}
Learned world models predict future environment states from observations, ranging from latent dynamics models~\citep{ha2018recurrent,hafner2019learning} to feature-space predictive approaches such as I-JEPA and DINO-WM~\citep{assran2023self,zhou2024dino}. Recent work adopts explicit 3D representations for embodied prediction: PointWorld~\citep{Huang2026PointWorld} models action-conditioned scene evolution with 3D point flows, while RoboFlow4D~\citep{lin2026roboflow4d} predicts multi-frame 3D motion fields for robotic manipulation. Despite this progress, rigid-body contact remains challenging because interactions can induce abrupt changes in global object motion, resulting in discontinuous dynamics that are difficult for smooth learned predictors to capture~\citep{allen2022learning}.

\textbf{Geometric Representations for Rigid-Body Simulation.}
Early learned physics models introduced inductive biases for predicting physical interactions~\citep{battaglia2016interaction, sanchez2018graph}, which were later extended to particle-based simulators for objects and fluids~\citep{li2018learning, sanchez2020learning}. Mesh-based approaches further exploit explicit geometric structure by performing message passing over simulation meshes~\citep{pfaff2020learning}. For rigid-body contact dynamics, FIGNet~\citep{allen2022learning} constructs interactions between mesh faces, preserving local surface geometry around contacts. Recent mesh-based models have explored hierarchical structures for propagating collision effects, while HOPNet~\citep{Wei2025integrating} explicitly represents vertices, edges, mesh triangles, contacts, and objects through higher-order topological relations. To relax the dependence on meshes, SDF-Sim~\citep{rubanova2024learning} represents object geometry using learned signed-distance functions, whereas RigidFormer~\citep{dou2026rigidformer} removes mesh connectivity entirely and operates on point-cloud inputs. Our method follows the mesh-free setting, while retaining point-level geometry for sparse local contact reasoning rather than compressing interactions primarily into coarse object-level representations.

\vspace{-0.15in}
\section{Method}\label{sec:Method}
\vspace{-0.15in}
In this section, we formulate the problem of modeling rigid-body contact dynamics and introduce Rigid-body Contact Reasoning (\methodname), our neural simulator designed to solve the problem.

\textbf{Problem Formulation.} To model the rigid-body dynamics, we consider a scene containing $M$ rigid objects, where $M$ may vary across scenes. At time $t$, object $i$ is represented by $N_i$ sampled surface points $\mathbf{x}_t^i = [{x}_{t,1}^i,\ldots,{x}_{t,N_i}^i]^\top
\in \mathbb{R}^{N_i\times 3}$,
with corresponding surface normals ${\nu}_t^i$, center of mass
$c_t^i$, and physical properties
$\boldsymbol{\phi}^i=[(m^i)^{-1},\mu^i,e^i]$.
We denote the full scene state by
$\mathbf{s}_t = \left\{\boldsymbol{x}_t^i,{\nu}_t^i,c_t^i\right\}_{i=1}^{M}$.
Given two consecutive states ${s}_{t-1}$ and ${s}_t$, together with
the static environment $\xi$, our goal is to build a  dynamics model $f_\theta$ (i.e., state transition model) to predict
$
    \hat{{s}}_{t+1}
    = f_\theta({s}_{t-1},{s}_t,\boldsymbol{\phi}, \xi).
$
Appendix~\ref{appendix:Problem Formulation} introduces more details.

\begin{figure}[t]
    \centering
    \includegraphics[width=\linewidth]{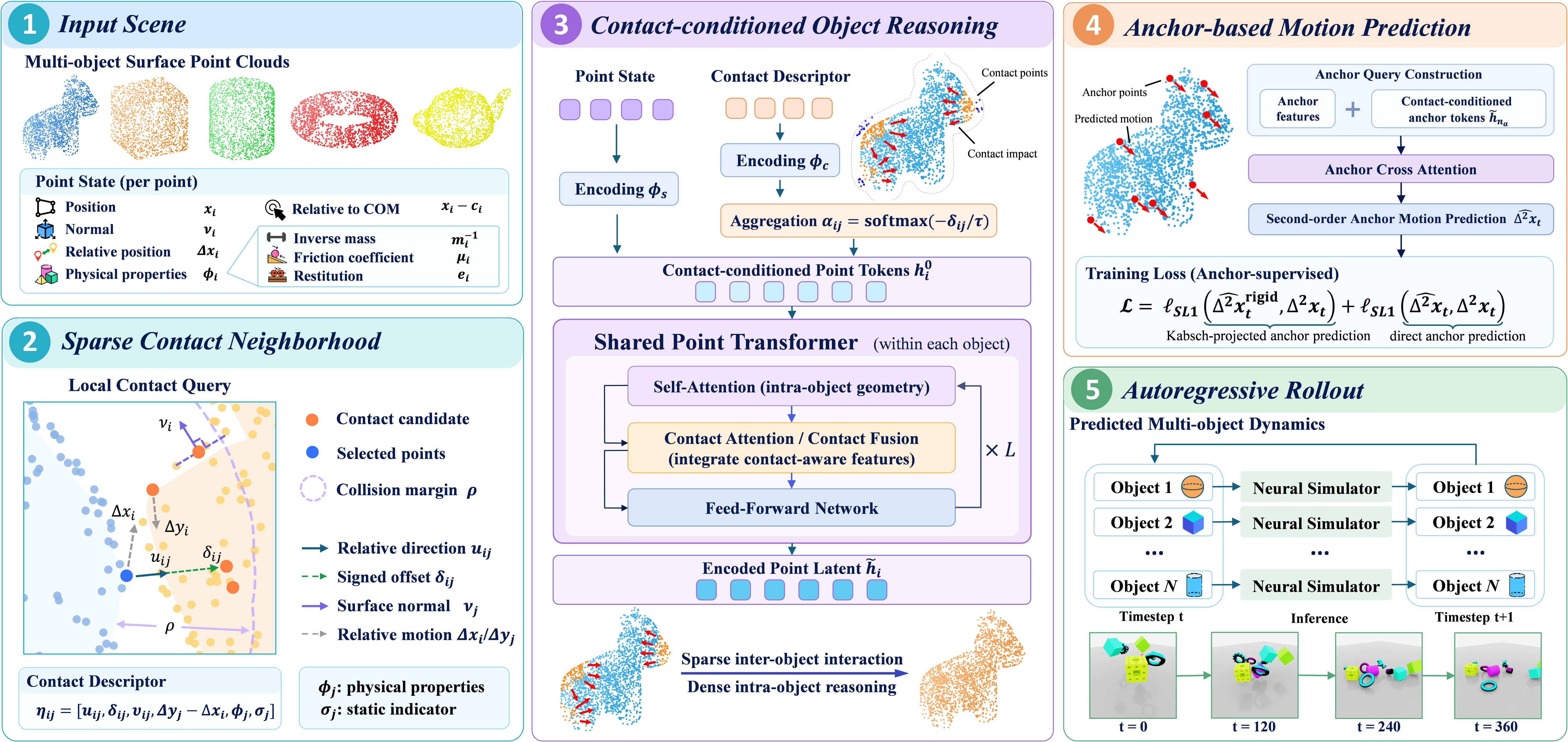}
    \caption{\textbf{Overview of RiCo.} From two consecutive scene states, RiCo constructs sparse local contact neighborhoods, fuses them with point-wise object states, and performs shared intra-object reasoning with a point transformer. The resulting contact-conditioned features are decoded at a small set of anchors to recover a rigid transformation for each object, and the predicted state is recursively used for autoregressive rollout.
    }
    \label{fig:method}
    \vspace{-0.2in}
\end{figure}

To effectively approximate $f_\theta$, \methodname\ models the state update in three stages: (1) constructing a sparse representation of local contacts, (2) reasoning over contacts within each object, and (3) predicting rigid-body motion from anchors. Figure~\ref{fig:method} provides an overview of the framework.

\vspace{-0.1in}
\subsection{Sparse Contact Neighborhood Representation}
\vspace{-0.1in}
\label{sec:local_contact}

\noindent\textbf{Contact neighborhood construction.}
Striving for concise representation of contact dynamics, we construct a sparse local interaction representation around each surface point, rather than modeling interactions over complete mesh-level or object-level representations~\citep{Wei2025integrating,dou2026rigidformer}. Specifically, we first perform an object-level broad-phase query over active object pairs using their surface-point AABBs, retaining only pairs whose minimum AABB separation is no greater than the contact radius $\rho$. 
More details can be found in appendix~\ref{app:contact_construction}.
Candidates from all retained objects and the static environment $\xi$ are then combined, from which the $K$ nearest candidates within $\rho$ are retained. Dynamic-object candidates therefore correspond to sampled surface points rather than closest points on continuous meshes, while static planar surfaces are queried analytically by projection.
We denote the candidate positions by $\{y_{t,n,k}^{i}\}_{k=1}^{K}$\footnote{If fewer than $K$ valid candidates exist, the remaining entries are padded and masked during subsequent aggregation; when no valid candidate exists, the entire contact neighborhood is masked and contributes zero contact information.}.

To characterize the potential contact response, we gather each candidate's surface normal $\widetilde{\nu}_{t,n,k}^{i}$, finite-difference displacement $\Delta y_{t,n,k}^{i}$, and physical properties $\widetilde{\phi}_{t,n,k}^{i}$, 
comprising the inverse mass, friction coefficient, and coefficient of restitution of the corresponding external object or surface. Additionally, a binary indicator $\sigma_{t,n,k}^{i}$ distinguishes dynamic geometry ($\sigma=0$) from static objects ($\sigma=1$), with static candidates assigned zero displacement and zero inverse mass.
We combine these quantities with the relative geometry and motion between the source point $x_{t,n}^{i}$ and each external candidate $y_{t,n,k}^{i}$ to form a 14-dimensional interaction descriptor:
\begin{equation}
\begin{aligned}
&\eta_{t,n,k}^{i} = \left[u_{t,n,k}^{i},\; \delta_{t,n,k}^{i},\; \widetilde{\nu}_{t,n,k}^{i},\; \Delta y_{t,n,k}^{i}-\Delta x_{t,n}^{i},\; \widetilde{\phi}_{t,n,k}^{i},\; \sigma_{t,n,k}^{i}\right] \in \mathbb{R}^{14}, 
\end{aligned}
\label{eq:contact_descriptor}
\end{equation}
where$\;u_{t,n,k}^{i} = \frac{y_{t,n,k}^{i}-x_{t,n}^{i}}{\left\|y_{t,n,k}^{i}-x_{t,n}^{i}\right\|_{2}+\varepsilon},\;
\delta_{t,n,k}^{i} = \left(x_{t,n}^{i}-y_{t,n,k}^{i}\right)^{\top}\widetilde{\nu}_{t,n,k}^{i}. $Within this formulation, $u_{t,n,k}^{i}$ encodes the direction from the source point to the candidate, with $\varepsilon>0$ ensuring numerical stability. The relative displacement $\Delta y_{t,n,k}^{i}-\Delta x_{t,n}^{i}$ captures approach or separation and tangential sliding, while $\widetilde{\phi}_{t,n,k}^{i}$ specifies the physical properties of the external object. We also introduce an estimated signed distance $\delta_{t,n,k}^{i}$, which measures the signed offset of the source point from the local tangent plane on the candidate surface. Unlike the unsigned distance, its sign explicitly distinguishes the two sides of the local surface, providing a simple geometric cue to identify the potential penetration.

\noindent\textbf{Point State Representation.}
To model how the source object responds to local interactions, we complement the contact descriptors with a point-wise state representation of its own geometry, motion, and physical properties.
For point $n$ on object $i$, we define this state as
$
s_{t,n}^{i} = \left[x_{t,n}^{i}-c_{t}^{i},\; \Delta x_{t,n}^{i},\; \nu_{t,n}^{i},\; \phi^{i},\; x_{t,n}^{i}-x_{0,n}^{i}\right] \in \mathbb{R}^{15}.
\label{eq:point_state}
$
Within this representation, $x_{t,n}^{i}-c_{t}^{i}$ locates the point relative to the object's center of mass.
The finite-difference displacement $\Delta x_{t,n}^{i}=x_{t,n}^{i}-x_{t-1,n}^{i}$ captures its motion over the preceding time step, while $\nu_{t,n}^{i}$ describes the local surface normal.
The object-level properties $\phi^{i}=[(m^i)^{-1},\mu^i,e^i]$ specify the inverse mass, friction coefficient, and restitution coefficient that govern contact response.
Finally, $x_{t,n}^{i}-x_{0,n}^{i}$ records the point's displacement from its initial position, encoding its accumulated motion relative to the reference configuration.

Together, the point state $s_{t,n}^{i}$ and the local interaction descriptors $\{\eta_{t,n,k}^{i}\}_{k=1}^{K}$ capture the source point's own geometry, motion, and physical properties alongside those of nearby external surfaces.
\vspace{-0.1in}
\subsection{Intra-Object Contact Reasoning}
\label{sec:object_dynamics}
\vspace{-0.1in}

While contact neighborhoods identify potential interactions between nearby surfaces, predicting the resulting rigid-body motion requires reasoning jointly about contacts across the object's surface.
We therefore fuse local contact descriptors with each point's state to form contact-conditioned tokens, then use a shared point Transformer to integrate these tokens within each object.

\noindent\textbf{Contact Feature Aggregation.}
We embed the point state $s_{t,n}^{i}$ and each local interaction descriptor $\eta_{t,n,k}^{i}$ using MLPs $\phi_s$ and $\phi_c$, respectively.
To prioritize nearby surfaces while retaining information from multiple contact candidates, we aggregate the interaction embeddings {$\phi_c(\eta_{t,n,k}^{i})$} using distance-dependent weights and fuse them with the point state embedding {$\phi_s(s_{t,n}^{i})$}.
For the valid candidate set $\mathcal{C}_{t,n}^{i}\subseteq\{1,\ldots,K\}$, we define
\begin{align}
&h_{t,n}^{i,0} = \operatorname{RMSNorm}\!\left(\phi_s(s_{t,n}^{i})+\sum_{k\in\mathcal{C}_{t,n}^{i}}\alpha_{t,n,k}^{i}\phi_c(\eta_{t,n,k}^{i})\right),\label{eq:contact_fusion}
\end{align}
where$\; \alpha_{t,n,k}^{i} = \frac{\exp(-d_{t,n,k}^{i}/\tau)}{\sum_{l\in\mathcal{C}_{t,n}^{i}}\exp(-d_{t,n,l}^{i}/\tau)}, \;\text{and}\; d_{t,n,k}^{i} = \|\mathbf{y}_{t,n,k}^{i}-\mathbf{x}_{t,n}^{i}\|_2\nonumber.$ Here, $d_{t,n,k}^{i}$ is the Euclidean distance, and $\tau>0$ is a learnable temperature controlling the concentration of the weights \footnote{The weights are normalized over valid candidates only, while points with no valid candidate receive a zero contact embedding. The fused representation is then normalized using RMSNorm, yielding $h_{t,n}^{i,0}$ for subsequent intra-object reasoning.}.

\noindent\textbf{Object-Wide Contact Integration.}
Predicting an object's response to contact requires reasoning about its interactions with surrounding objects and static surfaces.
Although these interactions are encoded locally, their combined effect on rigid-body motion depends on contact conditions across the object's surface.
As a result, for each object $i$, we collect the contact-conditioned point tokens as $\boldsymbol{h}_t^{i,0}=[h_{t,1}^{i,0},\ldots,h_{t,N}^{i,0}]$ and process them independently using a shared $B$-block point Transformer $\mathcal{T}_\theta$. Each block performs gated multi-head self-attention followed by a feed-forward network. For a single attention head in block $b$, we compute:
\begin{equation}
\begin{aligned}
\operatorname{Attn}_{\theta_b}(\boldsymbol{h}_t^{i,b})
&=
\left[\boldsymbol{g}_t^{i,b}\odot
\operatorname{softmax}\!\left(
\frac{(\boldsymbol{h}_t^{i,b}W_Q^b)(\boldsymbol{h}_t^{i,b}W_K^b)^\top}{\sqrt{d_{\mathrm{attn}}}}
\right)\boldsymbol{h}_t^{i,b}W_V^b\right]W_O^b, \\
\end{aligned}
\label{eq:intra_object_attention}
\end{equation}
where $\boldsymbol{h}_t^{i,b}$ denotes the point tokens of object $i$ at block $b$, and $\boldsymbol{g}_t^{i,b}=\sigma\!\left(\boldsymbol{h}_t^{i,b}W_G^b\right)$ is a learned sigmoid gate that modulates the attended features before output projection. $W_Q^b$, $W_K^b$, $W_V^b$, $W_G^b$, and $W_O^b$ denote the query, key, value, gate, and output projections, respectively, and $d_{\mathrm{attn}}$ is the query/key dimension of each head. After $B$ blocks, the resulting point representations are denoted by $\widetilde{\boldsymbol{h}}_t^i=\mathcal{T}_\theta(\boldsymbol{h}_t^{i,0})$. The Transformer parameters are shared across objects, while all queries, keys, and values are derived exclusively from the point tokens of the same object.
Importantly, since information from external surfaces is incorporated through the local contact fusion in Eq.~\ref{eq:contact_fusion}, intra-object self-attention allows each point to access contact information from across its own object without dense cross-object attention.
By restricting cross-object information exchange to sparse local contact neighborhoods, our method avoids dense point-level interactions across the entire scene.
\vspace{-0.1in}
\subsection{Anchor-Based Rigid Motion Prediction and Training}
\label{sec:anchor_update}
\vspace{-0.1in}
\noindent\textbf{Anchor-based Motion Prediction.} As a contact-aware neural simulator, \methodname\ predicts each rigid object's future motion from its contact-conditioned tokens $H_t^i$.
Since all surface points move under the same rigid transformation, predicting their future positions separately is often unnecessary.
To preserve rigidity, we follow prior anchor-based formulations~\citep{dou2026rigidformer} and select $A$ anchor samples for each object using deterministic farthest point sampling (FPS)~\citep{qi2017pointnet++}. The corresponding sample indices $\{n_a^i\}_{a=1}^{A}$ remain fixed throughout the rollout, such that the anchors track the same surface samples over time. We predict motion updates only at these anchors and use them to recover the rigid transformation of the entire object.
\begin{equation}
\widehat{\Delta^2\boldsymbol{x}}_{t,\mathrm{anc}}^i = \mathcal{D}_{\omega}\!\left(\boldsymbol{z}_t^i,\widetilde{\boldsymbol{h}}_t^i\right)\;\text{where}\;z_{t,a}^{i} = \widetilde{h}_{t,n_a^i}^{i}+\phi_q\!\left(\left[x_{t,n_a^i}^{i}-c_t^i,\; \Delta x_{t,n_a^i}^{i}\right]\right)
\label{eq:anchor_dynamics}
\end{equation}
Here, $\boldsymbol{z}_t^i$ collects the $A$ anchor queries, and $\mathcal{D}_{\omega}$ denotes the anchor-to-point attention decoder, including its output projection to second-order anchor displacements.
By attending to all point features of its object, each anchor integrates contact information from across the surface to predict a motion update conditioned on its own position and recent displacement.

To advance the anchors from their current motion, we use the predicted second-order displacements $\widehat{\Delta^2\boldsymbol{x}}_{t,\mathrm{anc}}^i$ as learned corrections to a constant-displacement extrapolation, yielding unconstrained next positions through a Verlet-style update:
$
\widetilde{x}_{t+1,n_a^i}^{i} = 2x_{t,n_a^i}^{i}-x_{t-1,n_a^i}^{i}+\left[\widehat{\Delta^2\boldsymbol{x}}_{t,\mathrm{anc}}^i\right]_a
$
Here, $[\cdot]_a$ extracts the $a$-th row, which contains the predicted second-order displacement of the anchor indexed by $n_a^i$.
To recover a rigid transformation for the object, we align the current anchor positions with their predicted positions using the Kabsch algorithm~\citep{kabsch1976solution} to obtain the recoverd rotation and translation $(\widehat{R}_t^i,\widehat{b}_t^i)$.
We apply them to every surface point and the center of mass, while updating surface normals using only the rotation:
$
\widehat{x}_{t+1,n}^i = \widehat{R}_t^i x_{t,n}^i+\widehat{b}_t^i$, $
\widehat{\nu}_{t+1,n}^i = \widehat{R}_t^i\nu_{t,n}^i$, 
$\widehat{c}_{t+1}^i = \widehat{R}_t^i c_t^i+\widehat{b}_t^i
$. 

\noindent\textbf{Anchor-Supervised Motion Training.} 
Applying the recovered rigid transformation to the current anchors yields $\widehat{x}_{t+1,n_a^i}^{i}$.
For the fixed anchor indices $\{n_a^i\}_{a=1}^{A}$, we compute the ground-truth and rigid-projected second-order displacements, respectively, as:
\begin{equation}
\left[\Delta^2\boldsymbol{x}_{t,\mathrm{anc}}^{i,*}\right]_a = x_{t+1,n_a^i}^{i,*}-2x_{t,n_a^i}^{i,*}+x_{t-1,n_a^i}^{i,*}, \qquad
\left[\widehat{\Delta^2\boldsymbol{x}}_{t,\mathrm{anc}}^{i,\mathrm{rigid}}\right]_a = \widehat{x}_{t+1,n_a^i}^{i}-2x_{t,n_a^i}^{i,*}+x_{t-1,n_a^i}^{i,*}.
\label{eq:anchor_displacement_targets}
\end{equation}
Here, $*$ denotes ground-truth quantities and $[\cdot]_a$ extracts the $a$-th row.
We supervise both the direct prediction and its rigid-projected counterpart in the following objective:
\begin{equation}
\mathcal{L} = \ell_{\mathrm{SL1}}\!\left(\widehat{\Delta^2\boldsymbol{x}}_{t,\mathrm{anc}}^i,\Delta^2\boldsymbol{x}_{t,\mathrm{anc}}^{i,*}\right) + \ell_{\mathrm{SL1}}\!\left(\widehat{\Delta^2\boldsymbol{x}}_{t,\mathrm{anc}}^{i,\mathrm{rigid}},\Delta^2\boldsymbol{x}_{t,\mathrm{anc}}^{i,*}\right).
\label{eq:training_objective}
\end{equation}
where $\ell_{\mathrm{SL1}}$ denotes the smooth-L1 loss. Both loss terms are computed in the normalized target space. Further implementation details are provided in Appendix~\ref{app:surface_representation}.
\vspace{-0.15in}
\section{Experiment}
\label{experiment}
\vspace{-0.15in}
\textbf{Datasets \& Metrics.}
We evaluate on MOVi-Sphere, MOVi-A, and MOVi-B~\citep{greff2022kubric}. The datasets span increasing geometric diversity, from spheres to primitive shapes and more complex object geometries. We report autoregressive position RMSE (m) and orientation RMSE (degrees) at 50, 75, and 100 frames~\citep{allen2022learning, rubanova2024learning, Wei2025integrating}. We additionally measure excess penetration frequency and mean penetration depth relative to the ground-truth trajectories in each dataset. More details of metrics can be found in appendix~\ref{app:metric_details}. The best results are highlighted in bold, while
the second-best results are indicated by underlined italics.

\textbf{Baselines.}
We compare \methodname\ with various baselines:
mesh-based simulators (e.g. FIGNet, HOPNet), point-based methods (VPD~\citep{whitney2024learning}), transformer-based approaches (HCMT~\citep{yu2024learning}) and point-based simulator (RigidFormer~\citep{dou2026rigidformer}).
For the main benchmark and cross-dataset generalization
experiments, we adopt the results reported by RigidFormer, including its reimplementations of FIGNet, HCMT, and VPD. For the remaining experiments, we focus on HOPNet~\citep{Wei2025integrating} and RigidFormer, two recent methods that are most relevant baselines of mesh-based and point-cloud-based methods, respectively. We use the officially released HOPNet checkpoint and our reimplementation of RigidFormer, as its official code is not published. 
\vspace{-0.1in}
\subsection{Motion Prediction Performance Across Rollout Horizons}
\vspace{-0.1in}
\label{main results}
We evaluate the autoregressive prediction accuracy of \methodname\ against various baselines across different object geometries and rollout horizons.

As shown in Table~\ref{tab:main-exp}, \methodname\ consistently achieves lower position and orientation errors than all baselines on MOVi-A and MOVi-B across all evaluated horizons.
On MOVi-A, compared with the strongest baseline, our model reduces position RMSE by approximately 35\% and orientation RMSE by 38--43\% over 50--100 prediction steps. Similarly, on MOVi-B, we obtain a 31--32\% reduction in position error and a 35--38\% reduction in orientation error.
On the geometrically simpler MOVi-Sphere benchmark, our translational accuracy is comparable to RigidFormer, differing by at most $0.001$\,m at each horizon, while consistently achieving lower orientation error. 
These results suggest that combining local contact representations with intra-object contact reasoning is particularly effective for predicting rigid-body interactions across diverse object shapes, while maintaining competitive performance in simpler spherical scenarios.

Notably, the prediction improvements persist throughout the evaluated autoregressive rollout horizons.
At 100 frames, \methodname\ achieves position and orientation RMSEs of $0.115$\,m/$11.40^\circ$ on MOVi-A and $0.111$\,m/$9.54^\circ$ on MOVi-B, respectively, substantially outperforming the corresponding baselines.
These results highlight the effectiveness of our contact-conditioned point Transformer, which integrates sparse local contact information through intra-object contact reasoning, for long-horizon rigid-body motion prediction.

\begin{table}[htbp]

\caption{
\textbf{Performance on MOVi-A, MOVi-B, and MOVi-Sphere.}
Each cell reports position RMSE (m) / orientation RMSE
($^\circ$) at prediction horizons of 50, 75, and 100 frames.
}

\label{tab:main-exp}

\centering

\setlength{\tabcolsep}{3pt}
\renewcommand{\arraystretch}{1.15}

\resizebox{\linewidth}{!}{%

\begin{tabular}{lccc ccc ccc}

\toprule

\textbf{Model}
& \multicolumn{3}{c}{\textbf{MOVi-A}}
& \multicolumn{3}{c}{\textbf{MOVi-B}}
& \multicolumn{3}{c}{\textbf{MOVi-Sphere}} \\

\cmidrule(lr){2-4}
\cmidrule(lr){5-7}
\cmidrule(lr){8-10}

& \textbf{50}
& \textbf{75}
& \textbf{100}

& \textbf{50}
& \textbf{75}
& \textbf{100}

& \textbf{50}
& \textbf{75}
& \textbf{100} \\

\midrule

FIGNet$_{\textit{reimpl}}$

& 0.132/7.10
& 0.285/14.62
& 0.492/23.30

& 0.141/7.39
& 0.300/15.16
& 0.516/24.96

& N/A
& N/A
& N/A \\

HCMT$_{\textit{reimpl}}$

& 0.239/5.70
& 0.538/11.82
& 0.951/18.40

& 0.237/4.72
& 0.527/9.80
& 0.932/17.43

& 0.243/4.19
& 0.541/8.21
& 0.956/13.81 \\

VPD$_{\textit{reimpl}}$

& 0.235/5.10
& 0.489/11.66
& 0.827/20.37

& 0.275/4.65
& 0.581/9.70
& 0.987/16.99

& 0.244/4.47
& 0.510/9.50
& 0.855/17.65 \\

HopNet

& 0.054/5.64
& 0.115/11.84
& 0.196/18.83

& \second{0.047}/4.91
& 0.101/10.35
& 0.176/17.91

& \second{0.034}/4.05
& 0.073/8.21
& 0.124/13.68 \\

RigidFormer

& \second{0.049}/\second{5.06}
& \second{0.103}/\second{10.90}
& \second{0.177}/\second{18.32}

& 0.050/\second{3.97}
& \second{0.095}/\second{8.51}
& \second{0.161}/\second{15.33}

& \best{0.026}/\second{3.00}
& \best{0.057}/\second{6.48}
& \best{0.099}/\second{11.19} \\

\midrule

\rowcolor{ricoblue}

\textbf{RiCo}

& \best{0.031}/\best{2.86}
& \best{0.067}/\best{6.49}
& \best{0.115}/\best{11.40}

& \best{0.032}/\best{2.59}
& \best{0.065}/\best{5.45}
& \best{0.111}/\best{9.54}

& \best{0.026}/\best{2.65}
& \second{0.058}/\best{5.59}
& \second{0.100}/\best{9.69} \\

\bottomrule

\end{tabular}

}

\vspace{1mm}

\begin{minipage}{\linewidth}
\footnotesize
\textit{Note:}
The subscript \textit{reimpl} denotes
reimplementations conducted by Rigidformer authors.
\end{minipage}
\vspace{-0.3in}
\end{table}
\vspace{-0.1in}
\subsection{Generalization Across Object Geometries}
\vspace{-0.1in}
A useful rigid-body dynamics model must predict interactions across diverse object geometries, including shapes that differ from those in its training data.
We therefore evaluate how well dynamics learned from one geometry distribution transfer to another.

As shown in Table~\ref{tab:generalization},
\methodname\ exhibits an asymmetric generalization
pattern, with particularly strong performance when
transferring from geometrically diverse training
distributions to simpler ones. When trained on the geometrically diverse MOVi-B dataset, our model transfers particularly well to its simpler constituent distributions. From MOVi-B to MOVi-Sphere, our 100-step position RMSE is $0.085$\,m, compared with $0.160$\,m for HopNet and $0.207$\,m for RigidFormer. Similarly, from MOVi-B to MOVi-A, our error is $0.112$\,m/$11.26^\circ$, substantially lower than both baselines. Notably, the additional geometry diversity in MOVi-B does not lead to negative transfer on simpler shapes. We observe similar improvements when transferring
from MOVi-A to MOVi-Sphere. In contrast, generalization from geometrically restricted training distributions to more diverse target distributions remains challenging. When transferring from MOVi-A to MOVi-B, \methodname\ achieves the lowest position error, although RigidFormer retains a slightly lower orientation error.
When trained exclusively on MOVi-Sphere, our method does not consistently outperform RigidFormer on the more geometrically diverse MOVi-A and MOVi-B benchmarks.

These results suggest that combining local contact representations with intra-object contact reasoning supports effective transfer across different geometry distributions, particularly when the training data provides sufficiently diverse geometric coverage. Nevertheless, generalization to more complex geometries remains dependent on the diversity of the training distribution.

\begin{table}[htbp]
\vspace{-0.05in}
\caption{
\textbf{Generalization performance across MOVi-Sphere (S),
MOVi-A (A), and MOVi-B (B).}
Each reports position RMSE (m) / orientation RMSE
($^\circ$) at prediction horizons of 75 and 100 frames.
}

\label{tab:generalization}

\centering
\setlength{\tabcolsep}{10pt}
\renewcommand{\arraystretch}{1.2}

\resizebox{\linewidth}{!}{%

\begin{tabular}{
cc
cc
cc
cc
}

\toprule

\textbf{Train} & \textbf{Test}
& \multicolumn{2}{c}{\textbf{HopNet}}
& \multicolumn{2}{c}{\textbf{RigidFormer}}
& \multicolumn{2}{c}{\textbf{RiCo}} \\

\cmidrule(lr){3-4}
\cmidrule(lr){5-6}
\cmidrule(lr){7-8}

&
& \textbf{75} & \textbf{100}
& \textbf{75} & \textbf{100}
& \textbf{75} & \textbf{100} \\

\midrule

% ==================================
% Train: MOVi-S
% ==================================

\multirow{2}{*}{\textbf{MOVi-S}}

& \textbf{MOVi-A}

& 0.112/11.29
& 0.197/\second{17.74}

& \best{0.107}/\second{11.27}
& \best{0.183}/17.92

& \second{0.110}/\best{10.15}
& \second{0.190}/\best{16.79} \\

& \textbf{MOVi-B}

& 0.106/9.75
& 0.188/17.13

& \best{0.096}/\best{9.42}
& \best{0.161}/\second{16.81}

& \second{0.103}/\second{9.63}
& \second{0.170}/\best{16.52} \\

\midrule

% ==================================
% Train: MOVi-A
% ==================================

\multirow{2}{*}{\textbf{MOVi-A}}

& \textbf{MOVi-S}

& 0.100/9.16
& 0.172/15.17

& \second{0.087}/\second{7.94}
& \second{0.153}/\second{14.27}

& \best{0.055}/\best{5.63}
& \best{0.094}/\best{9.83} \\

& \textbf{MOVi-B}

& \second{0.117}/10.77
& \second{0.202}/18.66

& 0.123/\best{9.29}
& 0.208/\best{17.08}

& \best{0.095}/\second{10.56}
& \best{0.164}/\second{17.93} \\

\midrule

% ==================================
% Train: MOVi-B
% ==================================

\multirow{2}{*}{\textbf{MOVi-B}}

& \textbf{MOVi-S}

& \second{0.095}/\second{9.09}
& \second{0.160}/\second{15.04}

& 0.123/10.52
& 0.207/17.88

& \best{0.049}/\best{5.31}
& \best{0.085}/\best{9.29} \\

& \textbf{MOVi-A}

& \second{0.120}/\second{12.08}
& \second{0.202}/\second{19.15}

& 0.140/12.25
& 0.234/20.28

& \best{0.066}/\best{6.38}
& \best{0.112}/\best{11.26} \\

\bottomrule

\end{tabular}%

}

\vspace{1mm}

\begin{minipage}{\linewidth}
\end{minipage}
\vspace{-0.1in}
\end{table}

\begin{figure}[ht]
    \centering

    \begin{subfigure}[t]{0.49\linewidth}
        \centering
        \includegraphics[width=\linewidth]{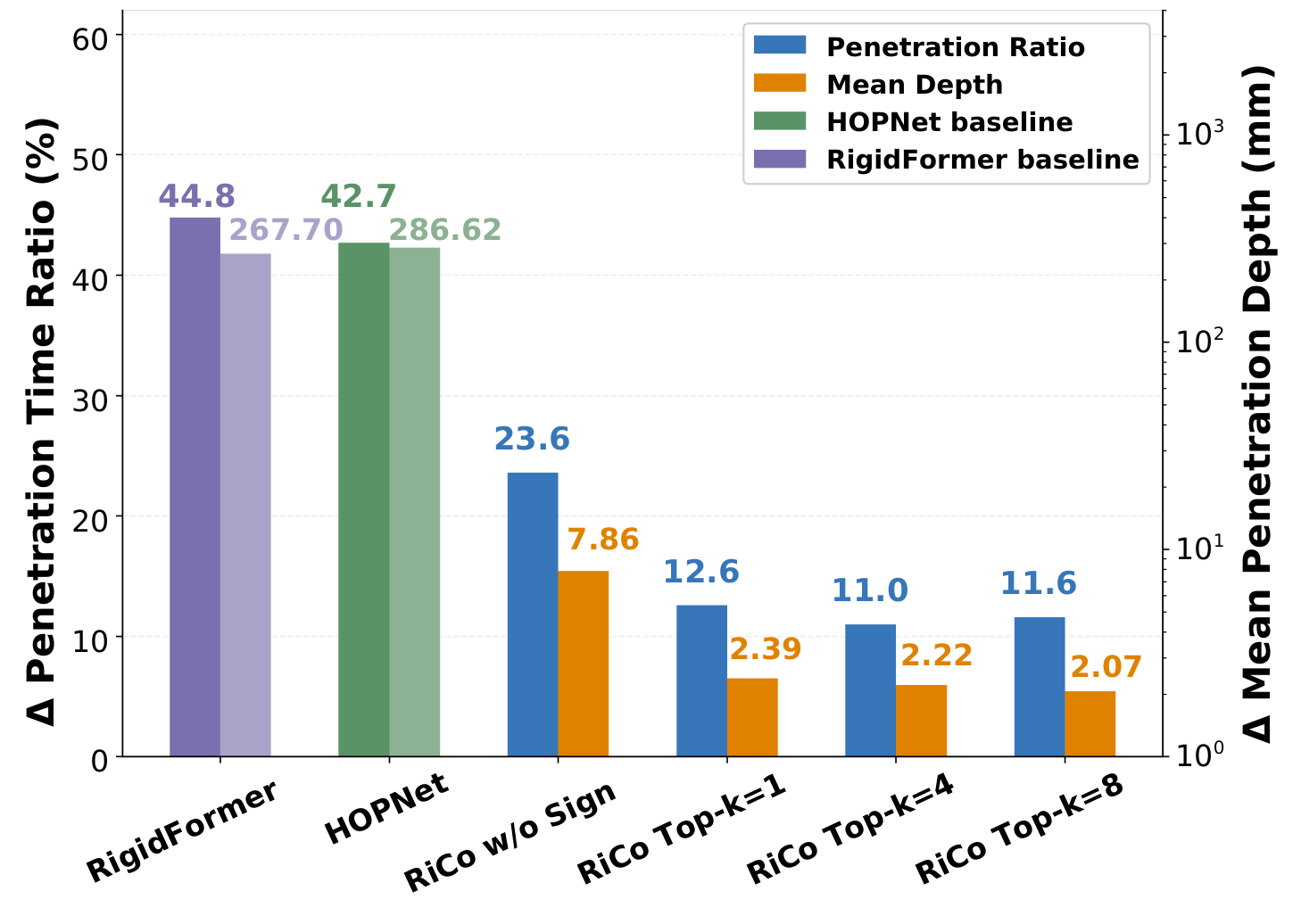}
        \caption{Relative penetration statistics.}
        \label{fig:local_contact_penetration}
    \end{subfigure}
    \hfill
    \begin{subfigure}[t]{0.49\linewidth}
        \centering
        \includegraphics[width=\linewidth]{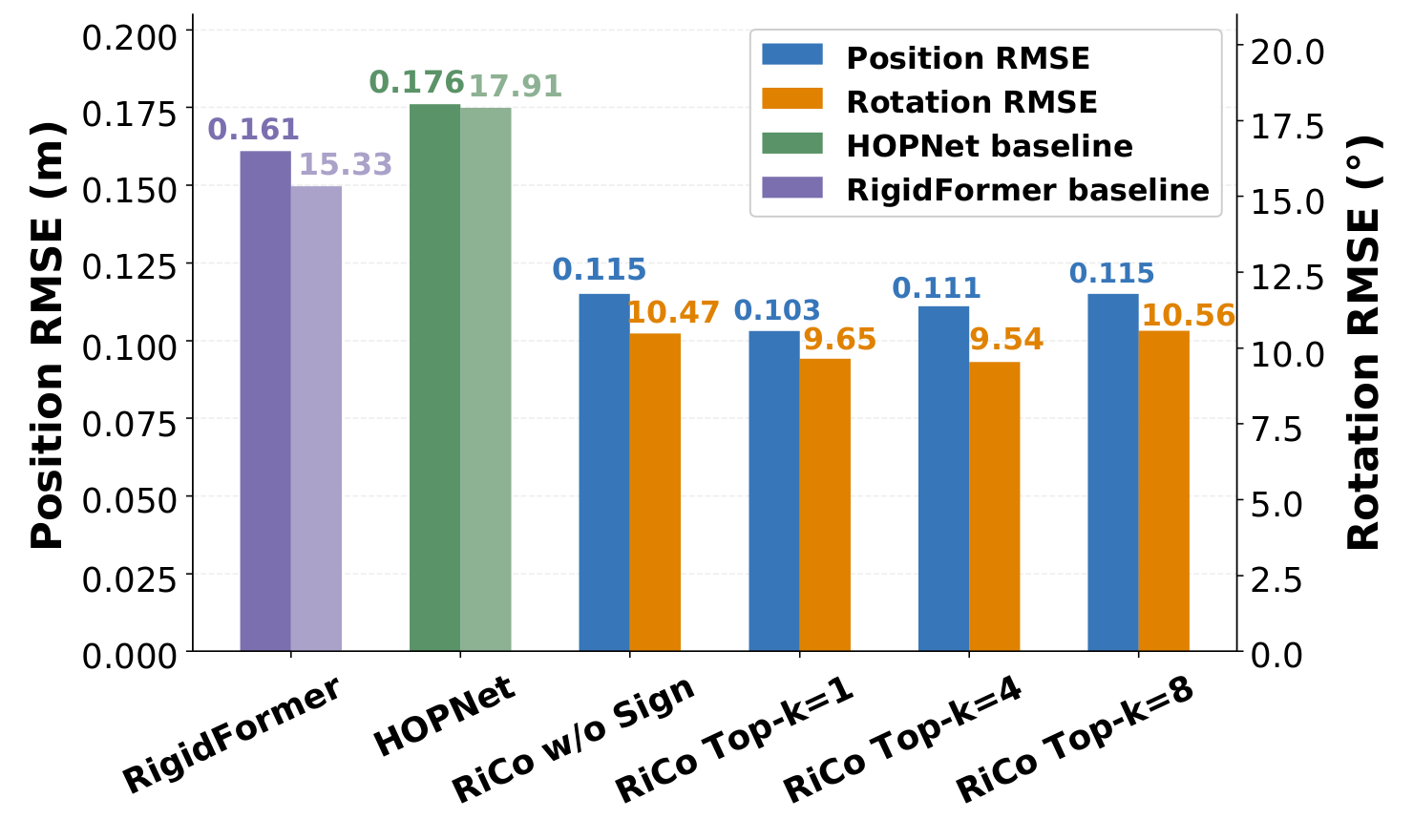}
        \caption{MOVi-B prediction error.}
        \label{fig:local_contact_rmse}
    \end{subfigure}
    
    \caption{
    \textbf{Ablation of local contact reasoning on MOVi-B.}
    (a) Ground-truth-relative penetration statistics.
    (b) Position and orientation RMSE at the
    100-step prediction horizon.
    For each method, the left bar shows the time ratio (left y-axis), and the right bar shows penetration depth (right y-axis).
    }
    \label{fig:local_contact_ablation}
    \vspace{-0.2in}
\end{figure}

\begin{figure}[htbp]
    \centering

    \begin{subfigure}[t]{0.49\linewidth}
        \centering
        \includegraphics[width=\linewidth]{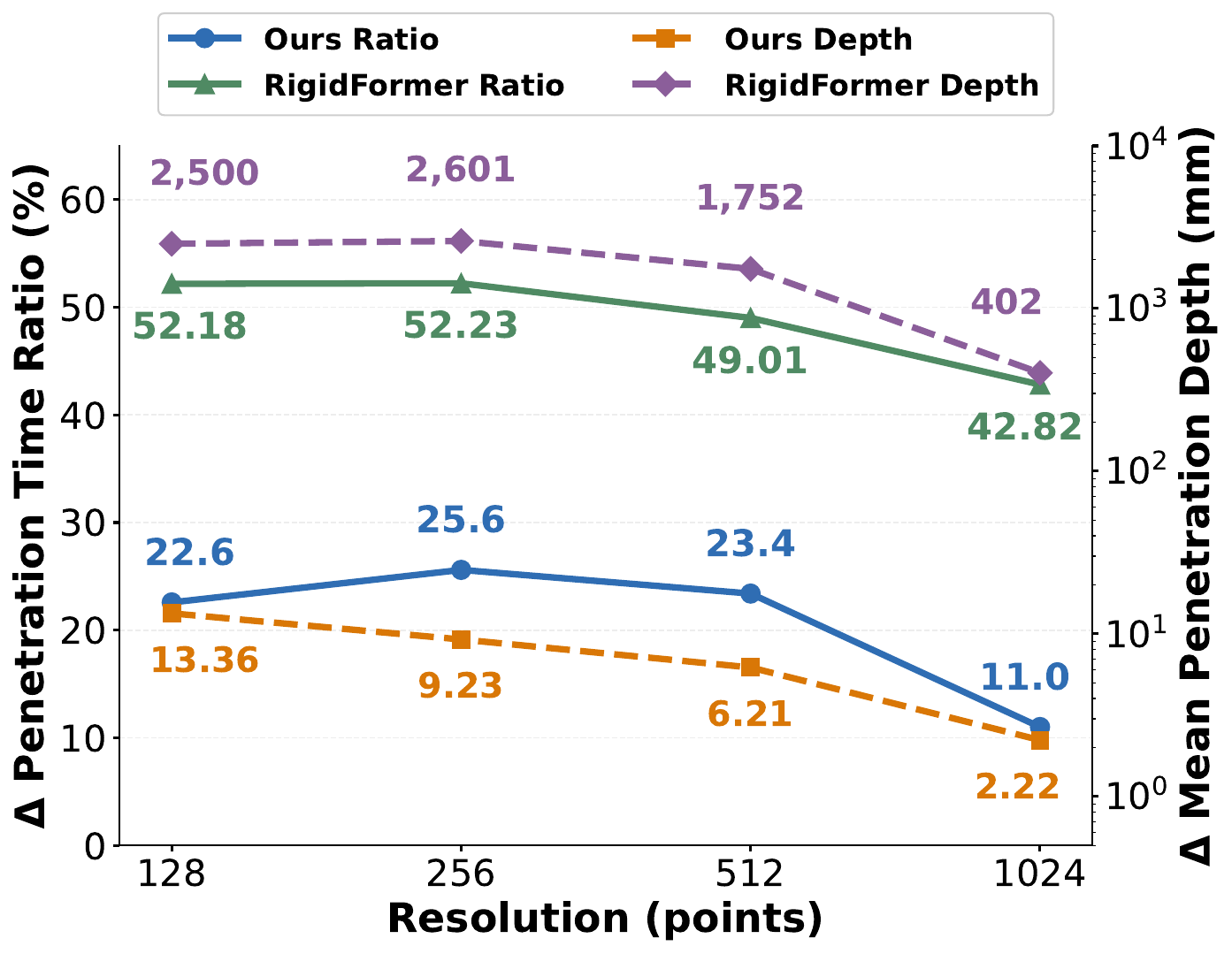}
        \caption{Relative penetration statistic.}
        \label{fig:resolution_penetration}
    \end{subfigure}
    \hfill
    \begin{subfigure}[t]{0.49\linewidth}
        \centering
        \includegraphics[width=\linewidth]{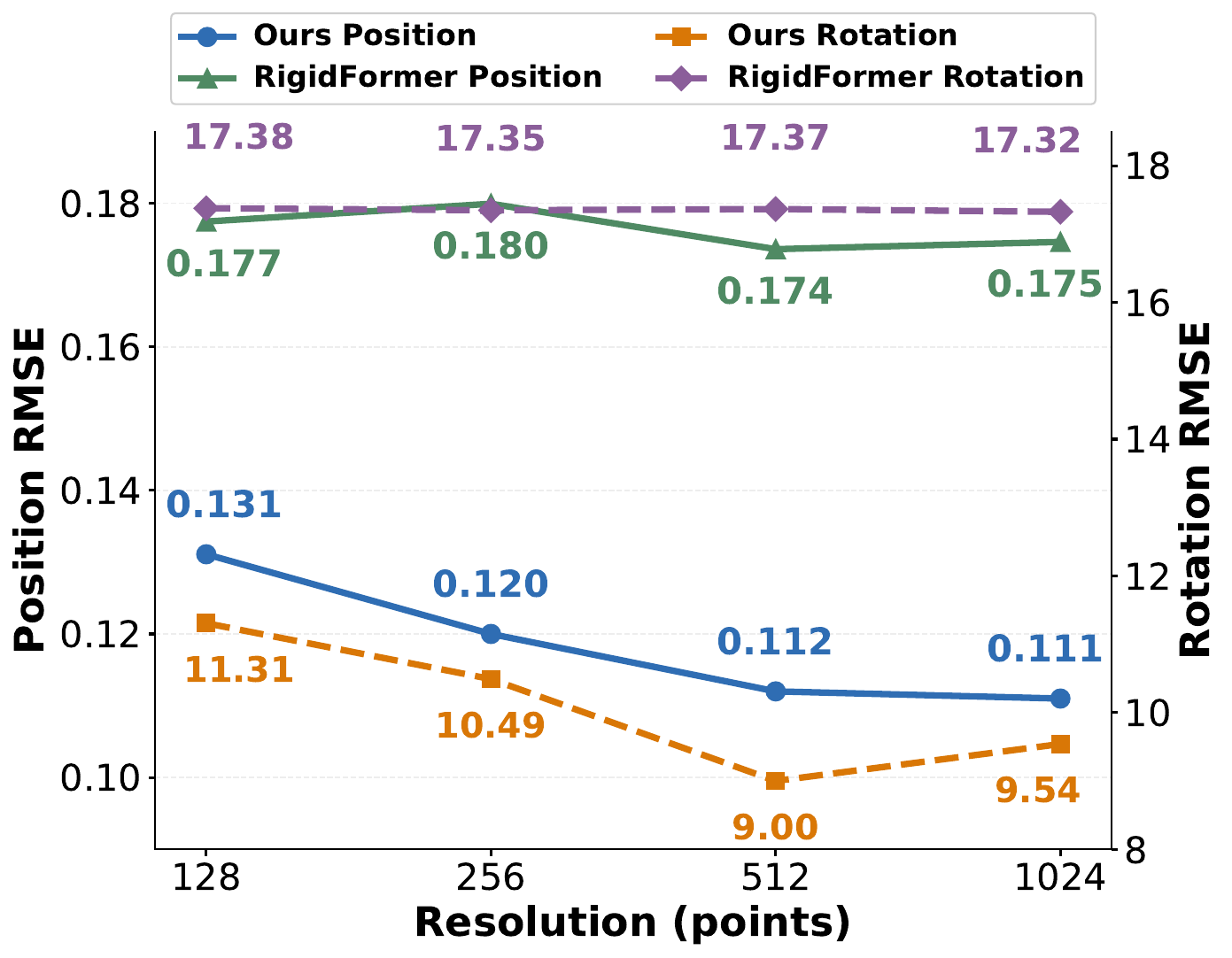}
        \caption{MOVi-B prediction error.}
        \label{fig:resolution_rmse}
    \end{subfigure}

    \caption{\textbf{Resolution and efficiency analysis on MOVi-B.} (a) Ground-truth-relative penetration statistics across different point-cloud resolutions. (b) Position and orientation prediction errors. For these curves, solid lines denote time ratios, measured on the left y-axis. Dashed lines denote penetration depths, measured on the right y-axis.
    }
    \label{fig:resolution_analysis}
    \vspace{-0.2in}
\end{figure}

\vspace{-0.1in}
\subsection{In-Depth Study on the Contact Fidelity}
\vspace{-0.1in}
\label{sec:ablation}

While trajectory RMSE measures prediction accuracy, it does not directly measure the physical consistency of predicted interactions. To evaluate contact fidelity, we introduce ground-truth-relative penetration metrics. It accounts for the geometric overlaps that already present in ground-truth trajectories. Specifically, we report the differences in penetration-time ratio ($\Delta$PTR) and mean penetration depth ($\Delta$MPD) between predicted and ground-truth trajectories. Detailed implementation can be seen in appendix~\ref{app:metric_details}. For these analyses, we evaluate RiCo against the released HOPNet checkpoint and our independent RigidFormer reimplementation using the same penetration evaluation procedure. We further investigate the effects of signed geometric information, contact neighborhood size, and point-cloud resolution on prediction accuracy and contact fidelity. We also conduct runtime analysis on point-cloud resolution in appendix~\ref{app:runtime}.

\textbf{Contact fidelity.}
As shown in Fig.~\ref{fig:local_contact_penetration}, \methodname\ achieves substantially lower ground-truth-relative penetration statistics than both baselines. On MOVi-B, \methodname\ records a $\Delta$PTR of $11.0\%$ and a $\Delta$MPD of $2.22$\,mm, compared with $42.7\%$ and $286.62$\,mm for HOPNet, and $44.8\%$ and $267.70$\,mm for our RigidFormer reimplementation. 
Together with the lower trajectory errors, these results indicate that \methodname\ improves prediction accuracy, while maintains high contact fidelity.

\textbf{Ablation of local contact representation.} 
As detailed in Sec.~\ref{sec:local_contact}, our local contact representation incorporates multiple geometric, motion, and physical features, most of which are either standard physical properties or have been explored in prior works, such as relative geometry and motion~\citep{Wei2025integrating,dou2026rigidformer}.
Therefore, we focus on evaluating the contribution of our estimated signed distance and investigating the effect of contact neighborhood size.

\emph{Estimated signed distance.} 
To investigate the importance of signed geometric information, we replace the estimated signed distance with its absolute value while retaining all other input features. As shown in Fig.~\ref{fig:local_contact_ablation}, this ablation only moderately affects trajectory RMSE but substantially degrades contact fidelity. Specifically, $\Delta$PTR increases from $11.0\%$ to $23.6\%$, while $\Delta$MPD increases from $2.22$ to $7.86$\,mm. These results suggest that explicitly encoding the signed geometric relationship between nearby surfaces helps the model better capture contact configurations and reduce interpenetration, even when its overall trajectory prediction accuracy changes only marginally.

\emph{Number of contact neighbors.}
We further vary the number of retained contact candidates $K \in \{1,4,8\}$. As shown in Fig.~\ref{fig:local_contact_ablation}, the model maintains relatively stable prediction accuracy and penetration statistics across different neighborhood sizes. While $K=8$ achieves slightly lower $\Delta$MPD, $K=4$ achieves the lowest $\Delta$PTR and orientation RMSE. 
These results suggest that a compact contact neighborhood is sufficient to achieve strong prediction performance under our benchmark.

\textbf{Exploiting high-resolution contact geometry.}
We next investigate whether increasing point-cloud resolution provides additional benefits for contact modeling, as shown in Fig.~\ref{fig:resolution_penetration} and \ref{fig:resolution_rmse}. Increasing \methodname's resolution from 512 to 1024 points yields only a marginal change in position RMSE, from $0.112$ to $0.111$\,m, but substantially improves contact fidelity. Specifically, $\Delta$PTR decreases from $23.4\%$ to $11.0\%$, while $\Delta$MPD decreases from $6.21$ to $2.22$\,mm.

In comparison, our RigidFormer reimplementation exhibits nearly unchanged trajectory RMSE across different resolutions, while its penetration statistics remain considerably higher despite improvements from increased surface sampling. These suggest that RiCo's local contact reasoning architecture is more effective in exploiting the richer contact information provided by high-resolution geometry. Compared with RigidFormer's shape compression and global interaction reasoning, RiCo retains fine-grained surface information through sparse local contact representations, supporting more accurate contact modeling and lower trajectory errors.

\vspace{-0.15in}
\subsection{Zero-shot Scalability to Large-scale Scenes}
\vspace{-0.1in}
\label{sec:scalability}

We further investigate whether the learned dynamics
can zero-shot generalize to large-scale scenes containing more interacting objects than those
encountered during training. Inspired by the large-scale evaluation setting of
SDF-Sim~\citep{rubanova2024learning},
we construct three large-scale scenarios, each
containing 270 rigid objects.
\methodname\ is trained exclusively on MOVi-B
scenes containing 3--10 objects and directly
evaluated on these large-scale scenarios.

\begin{wrapfigure}{r}{0.45\textwidth}
    \centering
    \vspace{-0.2in}
    % ===================================
    % Bottom: Independent Table
    % ===================================

    \captionof{table}{
        \textbf{Large-scale prediction.}
    }
    \label{tab:scalability}

    \vspace{0mm}

    \setlength{\tabcolsep}{3pt}
    \renewcommand{\arraystretch}{1.25}

    \resizebox{\linewidth}{!}{%
    \begin{tabular}{lcccc}
        \toprule

        & \multicolumn{3}{c}{
            \textbf{Position / Orientation RMSE}
        }
        & \textbf{Runtime} \\

        \cmidrule(lr){2-4}

        \textbf{Scene}
        & \textbf{120}
        & \textbf{240}
        & \textbf{360}
        & \textbf{FPS} \\

        \midrule

        Spheres
        & 0.029/1.73
        & 0.243/43.07
        & 0.713/89.90
        & 5.02 \\

        Spots
        & 0.019/1.21
        & 0.211/22.90
        & 0.785/86.74
        & 6.50 \\

        Knots
        & 0.017/0.81
        & 0.139/13.43
        & 0.843/78.00
        & 6.54 \\

        \bottomrule
    \end{tabular}%
    }
    \vspace{-0.1in}

\end{wrapfigure}

As shown in Table~\ref{tab:scalability}, we evaluate the autoregressive prediction accuracy at 120, 240 and 360 steps. 
Since collision events occur mainly during the latter part of these trajectories, we extend the evaluation horizon to 360 steps to further examine long-term prediction accuracy after extensive object interactions. More results can be found from appendix~\ref{app:scalability_temporal}. Compared with RigidFormer, which trains on large-scale WreckingBall scenes containing up to 217 objects, our evaluation directly applies a model trained on small-scale scenes to substantially larger configurations. Although RigidFormer reports a throughput of approximately 20 FPS on its large-scale benchmark, its experimental setting differs in object geometry and training settings.
\vspace{-0.15in}
\subsection{Transfer from Simulated to Real-World Dynamics}
\vspace{-0.1in}
To further investigate whether the learned local contact dynamics can generalize beyond simulated environments, we evaluate \methodname\ on real-world multi-ball collision experiments. In Figure~\ref{fig:real_world}, our real-world experiments are conducted on a billiard table with an effective playing area of $2.54 \times 1.27$\,m, using balls with a radius of $33.5$\,mm. We construct a corresponding simulated environment with comparable geometric configurations and train \methodname\ exclusively on simulated trajectories. The trained model is evaluated zero-shot on five real-world experimental sessions. Further details and qualitative results of the experimental setup and simulation environment are provided in the appendix~\ref{app:sim2real}. Quantitatively, \methodname\ achieves position RMSEs of $138.54$, $214.69$, and $272.42$\,mm at prediction horizons of 50, 75, and 100 frames. For reference, the physics engine achieves corresponding RMSEs of $65.37$, $147.84$, and $182.96$\,mm. We observe discrepancies when reconstructing real-world ball configurations in simulation using camera-estimated positions, suggesting that camera-based localization inaccuracy is a major source of error in our sim-to-real evaluation. These results provide preliminary evidence that the contact dynamics learned through sparse local geometric reasoning can transfer from simulation to real-world multi-object interactions. 
\vspace{-0.2in}
\section{Conclusion}
\vspace{-0.15in}
In this work, we introduce \methodname, a neural simulator that models rigid-body interactions through sparse local contact reasoning. By preserving fine-grained contact geometry while avoiding dense scene-level interactions, our approach achieves accurate and physically consistent predictions with high computational efficiency. Experiments demonstrate its ability to exploit high-resolution geometry, generalize across diverse object distributions, scale to scenes containing 270 objects and transfer to real world. Nevertheless, our current framework focuses on passive rigid-body dynamics and does not account for deformable objects or actuated systems. Extending local contact reasoning to these more complex physical interactions remains an important direction for future work.

\subsubsection*{Acknowledgments}
We are especially grateful to Wanxin Jin and Zhixian Xie from Arizona State University for their insightful discussions and valuable advice.

\bibliography{iclr2027_conference}
\bibliographystyle{iclr2027_conference}

\appendix

\section{Problem Formulation}\label{appendix:Problem Formulation}
\paragraph{Physical world modeling.}
A physical world model aims to capture how the physical environment evolves over time, enabling embodied agents to anticipate future states and reason about the consequences of their actions through predictive rollouts. Such predictive dynamics provide an important foundation for long-horizon reasoning, planning, and control, where an agent must evaluate possible future outcomes before acting. In physical environments, however, meaningful prediction requires not only modeling the motion of individual objects, but also reasoning about the interactions among them~\citep{sanchez2020learning}. This is particularly important in multi-object rigid-body systems, where contact and collision govern the transfer of momentum and can induce abrupt changes in both translational and rotational motion. Accurately modeling collision dynamics is therefore a key capability for physical world models to move beyond simple motion extrapolation and faithfully predict the long-horizon evolution of interacting scenes.

\paragraph{Rigid Body Contact Modeling.} 
In this work, we study the dynamics of contact between rigid bodies.
We consider a scene containing $M$ rigid objects.
At time $t$, object $i$ is represented by $N$ tracked surface samples, with positions $\boldsymbol{x}_t^i\in\mathbb{R}^{N\times 3}$ and surface normals $\boldsymbol{\nu}_t^i\in\mathbb{R}^{N\times 3}$.
We denote its center of mass by $c_t^i\in\mathbb{R}^3$ and its physical properties by $\phi^i=[(m^i)^{-1},\mu^i,e^i]$, where $m^i$, $\mu^i$, and $e^i$ are the mass, friction coefficient, and coefficient of restitution, respectively.

We collect the time-varying geometric quantities into the scene state $\mathbf{s}_t$ and the time-invariant physical properties into $\boldsymbol{\phi}$. In this way,
$\mathbf{s}_t = \left\{\boldsymbol{x}_t^i,\boldsymbol{\nu}_t^i,c_t^i\right\}_{i=1}^{M}$, and $\boldsymbol{\phi} = \left\{\phi^i\right\}_{i=1}^{M}$.

Surface-point identities are preserved over time, allowing us to compute their finite-difference displacements as $\Delta\boldsymbol{x}_t^i=\boldsymbol{x}_t^i-\boldsymbol{x}_{t-1}^i$.
We additionally retain the reference geometry $\boldsymbol{x}_0=\{\boldsymbol{x}_0^i\}_{i=1}^{M}$.
The dynamics model operates directly on these surface samples without requiring mesh connectivity.

Let $\xi$ denote the static environment together with its geometric and
physical properties. Given two consecutive scene states, we learn a transition
model
\begin{equation}
\widehat{\mathbf{s}}_{t+1}
=
f_{\theta}\!\left(
\mathbf{s}_{t},\mathbf{s}_{t-1},
\boldsymbol{\phi},
\xi
\right),
\label{eq:dynamics_transition}
\end{equation}
which predicts the next rigid-body configuration. At inference time, the same
transition model is applied recursively to generate long-horizon rollouts.

\section{Additional Method Details}
\label{app:method_details}

We provide implementation details of the surface-state representation,
sparse contact construction, network architecture, rigid-motion projection,
and training/inference procedures. Unless otherwise stated, the configurations
below describe our standard fixed-resolution local-contact implementation,
rather than separate resolution-ablation variants.

\subsection{Surface Representation and Feature Preprocessing}
\label{app:surface_representation}

\paragraph{Persistent surface representation.}
The offline dynamics model operates on object-aligned surface point
sets stored in trajectory data. Its default configuration represents
each active rigid body with $N=1024$ ordered surface points and
corresponding surface normals.

Point indices retain their physical correspondence across
consecutive frames; therefore, a point's finite-difference
displacement is obtained directly by subtracting its position
at the preceding frame, rather than by performing nearest-neighbor
matching across frames.

The standard MOVi training data uses $O_{\mathrm{data}}=10$
object slots per scene. An object-level activity mask identifies
occupied slots, and inactive slots are excluded from network
evaluation and losses. This fixed storage dimension is specific
to the training data and does not constrain the number of
objects supported by the model at inference time.

The default configuration does not use a variable-length
point mask within an active object: each occupied slot
contains precisely $1024$ point entries. The neural runtime
consumes these recorded point sets without resampling them
at each prediction step.

\paragraph{Object state and physical parameters.}
For a source point $\mathbf{x}_{t,n}^{i}$ on object $i$, we use the
physically defined center of mass $\mathbf{c}_t^i$ from the trajectory state,
not the arithmetic mean of the sampled surface points. The finite-difference
displacement is $\Delta\mathbf{x}_{t,n}^{i} =
\mathbf{x}_{t,n}^{i}-\mathbf{x}_{t-1,n}^{i}$; it is \emph{not}
divided by the simulation time step. Surface normals are normalized to unit
length before constructing the descriptors. Object-level physical parameters
$\boldsymbol\phi^i=[(m^i)^{-1},\mu^i,e^i]$ are broadcast to the points of that
object. We further preserve the initial surface configuration
$\mathbf{X}_0^i$ as a reference throughout the trajectory, so that each point
also encodes its displacement from its initial position.

Table~\ref{tab:app_input_features} specifies the exact order of the two
feature vectors. The source-point descriptor has $15$ channels, while each
external-contact descriptor has $14$ channels. The distance used for
neighbor weighting is maintained separately and is not duplicated as an
additional learned feature channel.

\begin{table}[htbp]
\centering
\caption{Input feature specification. All geometric positions and
finite-difference displacements are expressed in the same scene coordinate
system. Here $\mathbf{y}$ is an external candidate point,
$\mathbf{n}_{\mathrm{ext}}$ its unit normal, and
$\Delta\mathbf{y}$ its finite-difference displacement.}
\label{tab:app_input_features}
\small
\begin{tabular}{@{}llc@{}}
\toprule
Descriptor & Channels (in implementation order) & Dim. \\
\midrule
\multicolumn{3}{@{}l}{\emph{Source-point descriptor} $\mathbf{s}_{t,n}^i$} \\
& $\mathbf{x}_{t,n}^i-\mathbf{c}_t^i$ & 3 \\
& $\Delta\mathbf{x}_{t,n}^i$ & 3 \\
& $\boldsymbol\nu_{t,n}^i$ & 3 \\
& $(m^i)^{-1},\,\mu^i,\,e^i$ & 3 \\
& $\mathbf{x}_{t,n}^i-\mathbf{x}_{0,n}^i$ & 3 \\
\midrule
\multicolumn{3}{@{}l}{\emph{External-contact descriptor} $\boldsymbol\eta_{t,n,k}^i$} \\
& $(\mathbf{y}-\mathbf{x})/(\|\mathbf{y}-\mathbf{x}\|_2+\epsilon)$ & 3 \\
& $(\mathbf{x}-\mathbf{y})^\top\mathbf{n}_{\mathrm{ext}}$ & 1 \\
& $\mathbf{n}_{\mathrm{ext}}$ & 3 \\
& $\Delta\mathbf{y}-\Delta\mathbf{x}$ & 3 \\
& $m_{\mathrm{ext}}^{-1},\,\mu_{\mathrm{ext}},\,e_{\mathrm{ext}}$ & 3 \\
& $\sigma_{\mathrm{static}}\in\{0,1\}$ & 1 \\
\bottomrule
\end{tabular}
\end{table}

\paragraph{Feature and target normalization.}
The implementation uses three separate channel-wise normalizers: one for the
$15$-dimensional source features, one for the $14$-dimensional contact
features, and one for the $3$-dimensional second-order anchor-displacement
target. Each has the form
\begin{equation}
\label{eq:app_normalization}
\mathcal{N}(\mathbf{z})
=\frac{\mathbf{z}-\boldsymbol\mu}{\boldsymbol\sigma},
\qquad
\mathcal{N}^{-1}(\widehat{\mathbf{z}})
=\widehat{\mathbf{z}}\odot\boldsymbol\sigma+\boldsymbol\mu,
\end{equation}
where the division and multiplication act channel-wise. Normalization
statistics are fixed during model training and saved with the checkpoint;
they are not recomputed from validation or test trajectories. The provided
statistics-generation script accumulates moments over active objects in the
training split and computes contact statistics over valid contact entries
only. It uses a shared scale for each selected three-dimensional vector
group. In its physical-semantics policy, unit directions and normals,
friction/restitution coefficients, and the static indicator remain in their
original ranges, while the signed tangent-plane offset is divided by the
contact radius. Other continuous channels and the anchor target use
training-set statistics. An explicitly populated normalization configuration
takes precedence over a generated dataset sidecar; consequently, the exact
statistics used for a reported experiment are those saved in its checkpoint.

\subsection{Sparse Contact Neighborhood Construction}
\label{app:contact_construction}

\paragraph{Broad-phase object filtering.}
A contact neighborhood is recomputed from the current geometry at inference
time. For each active object $i$, we construct an axis-aligned bounding box
from its current surface points, with lower and upper corners
$\mathbf{l}_i$ and $\mathbf{u}_i$. For every pair of distinct active objects
$(i,j)$, their minimum box separation is
\begin{equation}
\label{eq:app_aabb}
d_{\mathrm{AABB}}(i,j)=
\left\|
\max\!\left(
\mathbf{l}_i-\mathbf{u}_j,
\mathbf{l}_j-\mathbf{u}_i,
\mathbf{0}
\right)
\right\|_2,
\end{equation}
with the maximum taken element-wise. Only pairs satisfying
$d_{\mathrm{AABB}}(i,j)\leq\rho$ are retained, where $\rho$ is the contact
radius. By default, we adopt $\rho=0.1$~m.
The pairwise geometric search is performed without gradient tracking.

\paragraph{Point-level search and global candidate selection.}
For each retained unordered pair $(i,j)$, the implementation computes a
pairwise Euclidean-distance matrix using the surface samples of the two
objects. In both directions, it retains at most $K$ nearest external
samples for every source point. The candidates obtained from all eligible
neighboring objects are then merged, and the closest $K$ dynamic candidates
are retained for each source point. Searches over eligible object pairs are
processed in bounded batches (\texttt{pair\_batch\_size}$=16$ in the default
configuration)

The default MOVi setting also includes a static horizontal plane at height
$z_{\mathrm{table}}$. For a source point
$\mathbf{x}=(x,y,z)$, its static candidate is the analytic projection
$\mathbf{y}_{\mathrm{table}}=(x,y,z_{\mathrm{table}})$, with normal
$(0,0,1)$, zero displacement, zero inverse mass, and the plane's assigned
friction and restitution. This analytic candidate is combined with the
$K$ retained dynamic candidates, and a final nearest-$K$ selection is
performed across both types. A selected candidate is valid only if it comes
from an eligible object or the plane and has a current Euclidean distance
no greater than $\rho$. Contacts with the source object's own points are
excluded. The default number of retained contact candidates is $K=4$.

\paragraph{Estimated signed offset and contact masking.}
For a selected candidate $\mathbf{y}$ with unit outward normal
$\mathbf{n}_{\mathrm{ext}}$, we compute
\begin{equation}
\label{eq:app_signed_offset}
\delta=(\mathbf{x}-\mathbf{y})^\top
\mathbf{n}_{\mathrm{ext}}.
\end{equation}
This quantity is the signed offset from the \emph{candidate's local tangent
plane}; it is an estimated local contact descriptor, not an exact
continuous-mesh signed-distance function. Its sign depends on the
orientation of the supplied surface normal. Together with the unit
direction, relative finite-difference displacement, external material
parameters, and static flag, it forms the $14$-dimensional descriptor
specified in Table~\ref{tab:app_input_features}. The feature builder
uses $\epsilon=10^{-8}$ in the direction normalization.

\paragraph{Offline contact caches versus online contact queries.}
The training dataset stores a final Top-$K$ target-index tensor at each
recorded frame. During teacher-forced training, the implementation reads
these target indices and reconstructs the relative contact features using
the recorded current point positions, displacements, normals, and physical
parameters; it does not redo nearest-neighbor search in every training
iteration. Invalid cached candidates are encoded with a sentinel index and
masked. By contrast, autoregressive inference invokes the live broad-phase
and nearest-neighbor query on the model's newly predicted geometry at every
step, allowing neighborhoods to change as contacts appear and disappear.
The contact-index selection itself is non-differentiable; the trainable
contact encoder and subsequent network operate on the selected descriptors.

\subsection{Network Architecture}
\label{app:network_architecture}

\paragraph{Pointwise encoders and distance-weighted fusion.}
The source-point encoder is a two-layer MLP
$15\rightarrow128\rightarrow384$ with a GELU activation. Each
$14$-dimensional contact descriptor is independently embedded by another
two-layer MLP $14\rightarrow128\rightarrow384$, using a Softplus
activation. The contact embeddings are aggregated by a masked,
distance-dependent softmax. In the implementation, the learnable
temperature is parameterized as
\begin{equation}
\label{eq:app_temperature}
\tau=\operatorname{softplus}(\tau_{\mathrm{raw}})+\tau_{\min},
\qquad \tau_{\min}=10^{-4},
\end{equation}
and initialized so that $\tau=0.005$. Softmax probabilities are computed
over valid candidates only and renormalized after masking; a source point
without any valid candidate receives a zero contact embedding. The sum of
the source and aggregated contact embeddings is then normalized with
RMSNorm to form the input point token.

\paragraph{Shared intra-object Transformer.}
Each active object's $N$ tokens are processed independently by the same
six-block point Transformer. Every block applies pre-normalized,
12-head self-attention followed by a separate pre-normalized feed-forward
network, with residual connections around both components. At token width
$d=384$, each attention head has dimension $32$. For the gated-attention
configuration, the implementation first computes standard scaled
dot-product attention and then multiplies the attended features by a
learned sigmoid gate computed from the normalized input tokens, before
the attention output projection. The feed-forward network uses a
$384\rightarrow3072$ projection followed by GLU, producing a hidden
width of $1536$, and a final $1536\rightarrow384$ projection. The
attention and feed-forward dropout settings are both zero in the default
configuration. No attention operation in this encoder directly connects
point tokens belonging to different objects.

\paragraph{Anchor-query decoder.}
The decoder creates one query for each of $A=8$ persistent anchor indices.
A query combines (i) the encoded token at that anchor and (ii) a geometry
embedding of the anchor's six-dimensional state, consisting of its
position relative to the physical center of mass and its preceding
finite-difference displacement. The geometry encoder is an MLP
$6\rightarrow128\rightarrow384$ with GELU. The resulting anchor queries
pass through six cross-attention blocks with 12 heads, attending to all
$N$ tokens of the \emph{same} object. Like the point Transformer, each
cross-attention block uses separate query/token RMSNorm, gated attention,
residual connections, and a GLU feed-forward network of hidden width
$1536$. An RMSNorm and an output MLP
$384\rightarrow128\rightarrow3$ with GELU produce the predicted
second-order displacement of every anchor. Decoder parameters are shared
across all objects.

\begin{table}[htbp]
\centering
\caption{Default local-contact model architecture. The two Transformer stacks operate
within each object; the indicated cross-attention is from anchor queries
to the object's own point tokens.}
\label{tab:app_architecture}
\small
\begin{tabular}{@{}ll@{}}
\toprule
Component & Configuration \\
\midrule
Maximum object slots / points per object & $10$ / $1024$ \\
Anchors / contact candidates per point & $8$ / $4$ \\
Source feature MLP & $15\!\to\!128\!\to\!384$, GELU \\
Contact feature MLP & $14\!\to\!128\!\to\!384$, Softplus \\
Contact pooling & Masked distance softmax \\
Contact temperature (initial/minimum) & $0.005$ / $10^{-4}$ \\
Feature fusion & Sum + RMSNorm \\
Point Transformer & $6$ blocks, $12$ heads \\
Anchor cross-attention decoder & $6$ blocks, $12$ heads \\
Token / per-head dimension & $384$ / $32$ \\
Transformer feed-forward hidden width & $1536$, GLU \\
Attention and FFN dropout & $0$ \\
Anchor geometry MLP & $6\!\to\!128\!\to\!384$, GELU \\
Output head & RMSNorm, $384\!\to\!128\!\to\!3$ \\
Kabsch numerical threshold & $10^{-6}$ \\
\bottomrule
\end{tabular}
\end{table}

\subsection{Anchor-Based Rigid Motion Prediction}
\label{app:rigid_projection}

\paragraph{Persistent anchor indices and displacement decoding.}
The trajectory interface provides eight fixed, distinct anchor indices for
each active object. These same indices are used when constructing the
anchor queries, supervising predictions, and aligning the predicted
configuration; the network does not dynamically reselect anchors in the
middle of a rollout. Let $\mathbf{q}_{t,a}^{i}$ be the current position
of anchor $a$ on object $i$. The decoder outputs a three-dimensional
\emph{second finite difference}, denoted by
$\widehat{\Delta^2\mathbf{q}}_{t,a}^{i}$, without division by the
physical time step squared. The unconstrained next anchor is
\begin{equation}
\label{eq:app_verlet}
\widetilde{\mathbf{q}}_{t+1,a}^{i}
=2\mathbf{q}_{t,a}^{i}-\mathbf{q}_{t-1,a}^{i}
+\widehat{\Delta^2\mathbf{q}}_{t,a}^{i}.
\end{equation}
The predicted normalized decoder output is converted back to physical
units by the target normalizer before this update.

\paragraph{Proper rigid alignment.}
We recover a single proper rigid transform independently for each active
object through unweighted least-squares alignment between its current
anchor positions and the unconstrained next positions. With column-vector
convention, let $\bar{\mathbf{q}}$ and
$\bar{\widetilde{\mathbf{q}}}$ be their respective centroids, and define
\begin{equation}
\label{eq:app_kabsch_svd}
\mathbf{H}=\sum_{a=1}^{A}
(\mathbf{q}_{t,a}-\bar{\mathbf{q}})
(\widetilde{\mathbf{q}}_{t+1,a}-
 \bar{\widetilde{\mathbf{q}}})^\top,
\qquad \mathbf{H}=\mathbf{U}\mathbf{\Sigma}\mathbf{V}^\top.
\end{equation}
To exclude reflections, the implementation computes
\begin{equation}
\label{eq:app_kabsch_solution}
\mathbf{R}=\mathbf{V}
\operatorname{diag}\!\left(
1,1,\operatorname{sign}\det(\mathbf{V}\mathbf{U}^\top)
\right)\mathbf{U}^\top,
\qquad
\mathbf{b}=\bar{\widetilde{\mathbf{q}}}
-\mathbf{R}\bar{\mathbf{q}}.
\end{equation}
The same $\mathrm{SE}(3)$ transform advances all object surface points
and its physical center of mass, while normals are updated by rotation
alone and renormalized to unit length. For inactive object slots, the
alignment implementation returns an identity rotation and zero
translation; their output states are subsequently masked. The Kabsch
implementation uses a numerical threshold of $\epsilon_{\rm K}=10^{-6}$
for masking and safe centroid computation.

\paragraph{Autoregressive Inference.}
At inference time, we initialize \methodname\ with two consecutive observed states at times $t-1$ and $t$ to predict the state at $t+1$.
At each step $t$, the recovered rotation $\widehat{R}_t^i$ and translation $\widehat{b}_t^i$ update the surface positions $\widehat{x}_{t+1,n}^i$ and center of mass $\widehat{c}_{t+1}^i$, while the rotation alone updates the surface normals $\widehat{\nu}_{t+1,n}^i$.
We then recompute the sparse local contact representation from the predicted geometry at $t+1$ for the next transition.

\section{Experimental Setup and Implementation}
\label{app:experimental_details}

We describe the offline data interface, optimization setup, evaluation
protocol, and metric implementations used in our local-contact experiments.
The architectural details of the standard model are provided in
Appendix~\ref{app:method_details}.

\subsection{Benchmarks and Data Processing}
\label{app:dataset_details}

\paragraph{Benchmark suites.}
We evaluate on MOVi-Sphere, MOVi-A, and MOVi-B using the publicly released datasets from HOPNet~\citep{Wei2025integrating}.
These benchmarks span different geometric distributions, enabling evaluations of in-distribution rollout prediction and cross-dataset generalization.
Our standard offline implementation consumes precomputed trajectories rather than rendering observations or resampling surface points during training.
We follow the original train/test partitions provided by HOPNet, with training and evaluation trajectories stored separately.

\paragraph{Trajectory representation.}
The standard MOVi training datasets are stored in HDF5 format,
using $O_{\mathrm{data}}=10$ object slots per scene,
$N=1024$ ordered surface points per active object,
and $A=8$ fixed anchor indices per object.
This fixed object dimension is specific to the offline
training data representation and does not impose
a limit on the number of objects supported by
\methodname.

For a trajectory with $T$ stored frames, the primary
state tensors have conceptual shapes
$[T,O_{\mathrm{data}},N,3]$ for surface positions
and normals,
$[T,O_{\mathrm{data}},3]$ for physical centers of mass,
and $[T,O_{\mathrm{data}}]$ for inverse mass,
friction, restitution, and object activity. Contact-target indices are stored as  $[T,O_{\mathrm{data}},N,K]$, where $K=4$ in the standard setting; these indices refer to neighboring surface samples or designated static contacts.

The dataset also stores trajectory lengths, terminal
flags, and persistent anchor indices.
The loader uses trajectory lengths to avoid padded
trailing frames and excludes training windows that
cross a terminal transition.
Its two-frame input and one-frame target configuration
produces three-frame training windows.

Since \methodname\ shares network parameters across
objects, its inference is not restricted to the
object count used during training.
This enables our zero-shot evaluation on scenes
containing 270 objects, as described in
Sec.~\ref{sec:scalability}.

\paragraph{Cached contacts during training.}
For a ground-truth training window, the loader reads the cached contact
target IDs corresponding to its current frame. Their positions, normals,
finite-difference displacements, and physical parameters are then gathered
from the current ground-truth state to reconstruct the contact descriptors.
This avoids repeating the discrete nearest-neighbor search for every
teacher-forced training instance. In contrast, autoregressive evaluation
reconstructs contact neighborhoods from the model's \emph{predicted}
current geometry at each new step. Consequently, the model does not use
future ground-truth contact assignments during inference. Additional
feature and target normalization details are given in
Appendix~\ref{app:surface_representation}.

\subsection{Optimization and Training Configuration}
\label{app:training_details}

\paragraph{Offline supervised training.}
The standard training loop receives two consecutive ground-truth states and predicts the next state with teacher forcing. The optimizer is AdamW. For each valid training object, supervision is applied to the direct and rigid-projected second-order anchor displacements using coordinate-wise smooth-$L1$ losses in normalized target space; both losses have unit weight. Objects contribute only if their slots are active at both the current and target frames. Loss numerators are summed over valid anchor coordinates.

\paragraph{Reference configuration.}
Table~\ref{tab:app_training_config} summarizes the reference offline training configuration. Configurations associated with separate benchmark training runs and ablations are stored alongside their corresponding checkpoints.

\begin{table}[t]
\centering
\caption{Reference offline optimization settings for the local-contact model.}
\label{tab:app_training_config}
\small
\begin{tabular}{@{}ll@{}}
\toprule
Setting & Value \\
\midrule
Optimizer & AdamW \\
Global batch size & $128$ \\
Per-device micro-batch size & $8$ \\
Context / target frames & $2$ / $1$ \\
Precision & \texttt{bf16} \\
Initial / final learning rate & $10^{-4}$ / $10^{-5}$ \\
Learning-rate schedule & Cosine decay \\
Weight decay & $0$ \\
Gradient-norm clipping & $1.0$ \\
Training epochs & $60$ \\
Autoregressive fine-tuning & Disabled \\
Direct / projected loss weights & $1.0$ / $1.0$ \\
Random seed & $0$ \\
\bottomrule
\end{tabular}
\end{table}

\paragraph{Implementation and model selection.}
The trainer supports gradient accumulation through micro-batches and
multi-GPU training through distributed data parallelism. Its learning-rate
schedule advances per optimizer update, and its \texttt{bf16} setting
applies automatic mixed precision to the neural forward computation.
Normalization parameters are computed from training data or loaded from
an explicitly specified configuration and are saved with the resulting
checkpoint. The available validation routine evaluates autoregressive
predictions at physical frame indices $50$, $75$, and $100$. These are
frame indices in the stored trajectory, not a claim that exactly that
many neural updates follow the two observation frames. The checkpoint selection criterion and computational resources are
specified for each final experiment.

\subsection{Evaluation Metrics and Measurement Procedures}
\label{app:metric_details}

\paragraph{Position and orientation errors.}
Let $\widehat{\mathbf c}_{r,t}^i$ and $\mathbf c_{r,t}^{i,*}$ be the
predicted and ground-truth physical centers of mass for object $i$ in
trajectory $r$ at evaluation frame $t$. Let $\mathcal{V}_{r,t}$ be the
set of objects active in both states. We first compute a per-scene
position error,
\begin{equation}
\label{eq:app_position_rmse}
E_{r,t}^{\mathrm{pos}} =
\sqrt{\frac{1}{|\mathcal V_{r,t}|}
\sum_{i\in\mathcal V_{r,t}}
\big\|\widehat{\mathbf c}_{r,t}^i-\mathbf c_{r,t}^{i,*}\big\|_2^2}.
\end{equation}
For orientation, we estimate the relative proper rotation
$\Delta\mathbf R_{r,t}^i\in\mathrm{SO}(3)$ between predicted and
ground-truth surface-point configurations using rigid alignment. Its
geodesic angle in degrees is
\begin{equation}
\label{eq:app_geodesic_angle}
\theta_{r,t}^i = \frac{180}{\pi}\arccos\!\left[
\operatorname{clip}\!\left(
\frac{\operatorname{tr}(\Delta\mathbf R_{r,t}^i)-1}{2},-1,1
\right)\right].
\end{equation}
The per-scene orientation error is then
\begin{equation}
\label{eq:app_orientation_rmse}
E_{r,t}^{\mathrm{rot}} =
\sqrt{\frac{1}{|\mathcal V_{r,t}|}
\sum_{i\in\mathcal V_{r,t}}(\theta_{r,t}^i)^2}.
\end{equation}
The reported frame-specific values are the arithmetic means of the
per-scene RMSEs across evaluated trajectories. Translation is expressed
in meters and orientation in degrees. This distinction matters because
averaging scene-level RMSEs is not generally equivalent to pooling all
objects across the dataset before taking a square root.

\paragraph{Collision geometry and penetration depth.}
Penetration is evaluated separately from the network's input-point
representation. The evaluation script loads the scaled collision meshes
associated with each object's shape, registers the mesh to the
frame-zero ground-truth surface points, and uses rigidly estimated poses
to place the mesh in every subsequent ground-truth or predicted frame.
For an object pair, an AABB check first rejects pairs with disjoint
boxes. For the remaining pairs, the script tests the vertices of each
mesh for occupancy inside the other mesh and queries the distances of
inside vertices to that mesh's surface. It takes the maximum measured
depth over the two query directions. Penetration of the horizontal
ground plane is evaluated from the minimum height of the transformed
mesh vertices. For each active object and frame, the recorded depth
$D_{r,t}^i\geq0$ is the maximum across all detected object--object
and object--ground violations; $P_{r,t}^i=\mathbb{1}[D_{r,t}^i>0]$
indicates a detected penetration. These measurements use a geometric
vertex-query approximation on the evaluation meshes, rather than an
exact continuous-surface intersection-depth solver. No additional
positive depth threshold is applied in the inspected script.

\begin{wrapfigure}{r}{0.45\columnwidth}
    \centering
    \vspace{-10pt}
    \includegraphics[width=\linewidth]{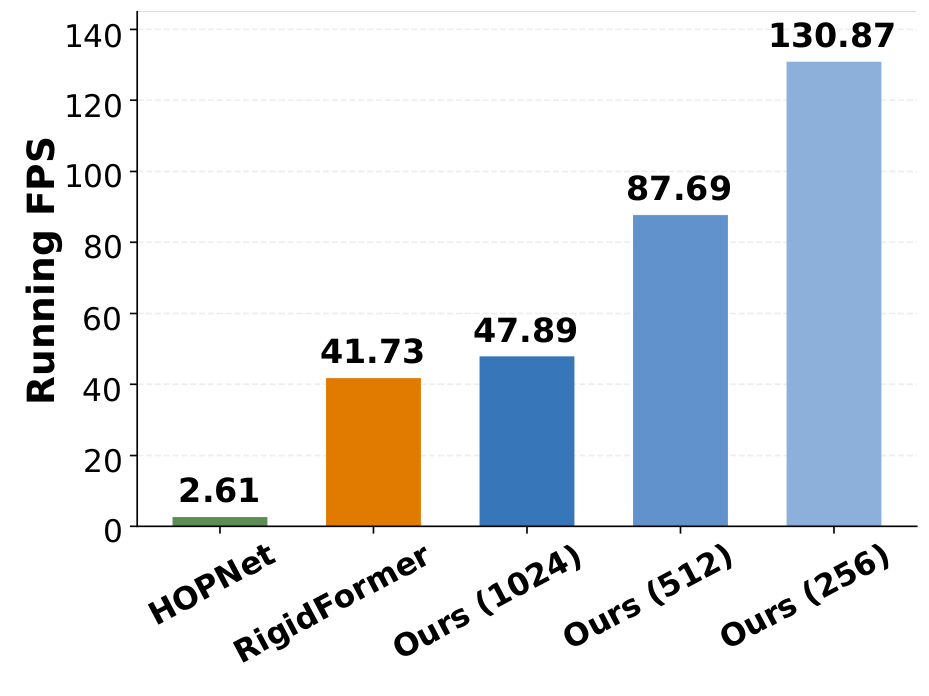}
    \vspace{-10pt}
    \caption{Runtime analysis.}
    \label{fig:resolution_runtime}
    \vspace{-10pt}
\end{wrapfigure}

\paragraph{Ground-truth-relative penetration statistics.}
Let $S$ denote either the predicted or ground-truth trajectories,
and let $a_{r,t}^i\in\{0,1\}$ indicate an active object.
The evaluated frame set $\mathcal T$ excludes the two initial frames.
Let $\mathcal Q$ contain all trajectory--object pairs active in at least
one evaluated frame. The penetration-time ratio implemented by our
evaluator is
\begin{equation}
\label{eq:app_ptr}
\operatorname{PTR}(S) =
\frac{1}{|\mathcal Q|}\sum_{(r,i)\in\mathcal Q}
\frac{\sum_{t\in\mathcal T}a_{r,t}^i P_{r,t}^{i,S}}
{\sum_{t\in\mathcal T}a_{r,t}^i}.
\end{equation}
Thus, the evaluator first computes the fraction of penetrating active
frames for each trajectory--object pair, then averages these fractions.
The mean penetration depth is computed over all \emph{detected
penetrating} active object--frame observations,
\begin{equation}
\label{eq:app_mpd}
\operatorname{MPD}(S)=
\frac{\sum_{r,i}\sum_{t\in\mathcal T}
a_{r,t}^i P_{r,t}^{i,S}D_{r,t}^{i,S}}
{\sum_{r,i}\sum_{t\in\mathcal T}a_{r,t}^i P_{r,t}^{i,S}},
\end{equation}
with MPD set to zero if the denominator vanishes. Depths are computed
in meters and converted to millimeters for display. Finally, the
\emph{ground-truth-relative} values printed by the current evaluation
script are the differences of these aggregated statistics,
\begin{equation}
\label{eq:app_relative_penetration}
\Delta\operatorname{PTR}=
\operatorname{PTR}(\mathrm{pred})-\operatorname{PTR}(\mathrm{GT}),
\qquad
\Delta\operatorname{MPD}=
\operatorname{MPD}(\mathrm{pred})-\operatorname{MPD}(\mathrm{GT}).
\end{equation}
The first quantity is reported in percentage points and the second in
millimeters. These are differences of dataset-level summaries; they are
\emph{not} the frequency or conditional depth of a separately defined
per-frame positive excess
$\max(0,D_{r,t}^{i,\mathrm{pred}}-D_{r,t}^{i,\mathrm{GT}})$.
This distinction is relevant when interpreting the physical-consistency
results in the main paper.

\subsection{Runtime analysis}
\label{app:runtime}

\noindent\textbf{Experimental setup.}
We evaluate computational efficiency on a single NVIDIA
GeForce RTX 5090 GPU using the same 120 test trajectories
for all methods. Runtime is measured with a batch size of
one over 100-step autoregressive rollouts. Each rollout is
initialized with two ground-truth frames, after which
subsequent states are predicted autoregressively without
accessing future ground-truth states.

We perform five warm-up iterations before timing and
synchronize CUDA execution immediately before and after
each measured rollout. Dataset loading, disk I/O,
visualization, and rendering are excluded from the
runtime measurements.

\textbf{Computational efficiency.}
We examine whether retaining high-resolution geometry introduces substantial computational overhead. As shown in Fig.~\ref{fig:resolution_runtime}, we evaluate baselines and \methodname\ on the same 120 test trajectories using a single NVIDIA RTX 5090 GPU. Despite operating on 1024 surface points per object, RiCo achieves $47.89$ FPS, slightly faster than our RigidFormer reimplementation ($41.73$ FPS) and over an order of magnitude faster than HOPNet ($2.61$ FPS). Reducing RiCo's resolution to 512 and 256 points further increases throughput to $87.69$ and $130.87$ FPS, respectively. These results demonstrate that sparse local contact reasoning enables high-resolution geometric modeling without sacrificing computational efficiency. 

\begin{wrapfigure}{r}{0.45\columnwidth}
    \vspace{-0.1in}
    \centering
    \includegraphics[width=\linewidth]{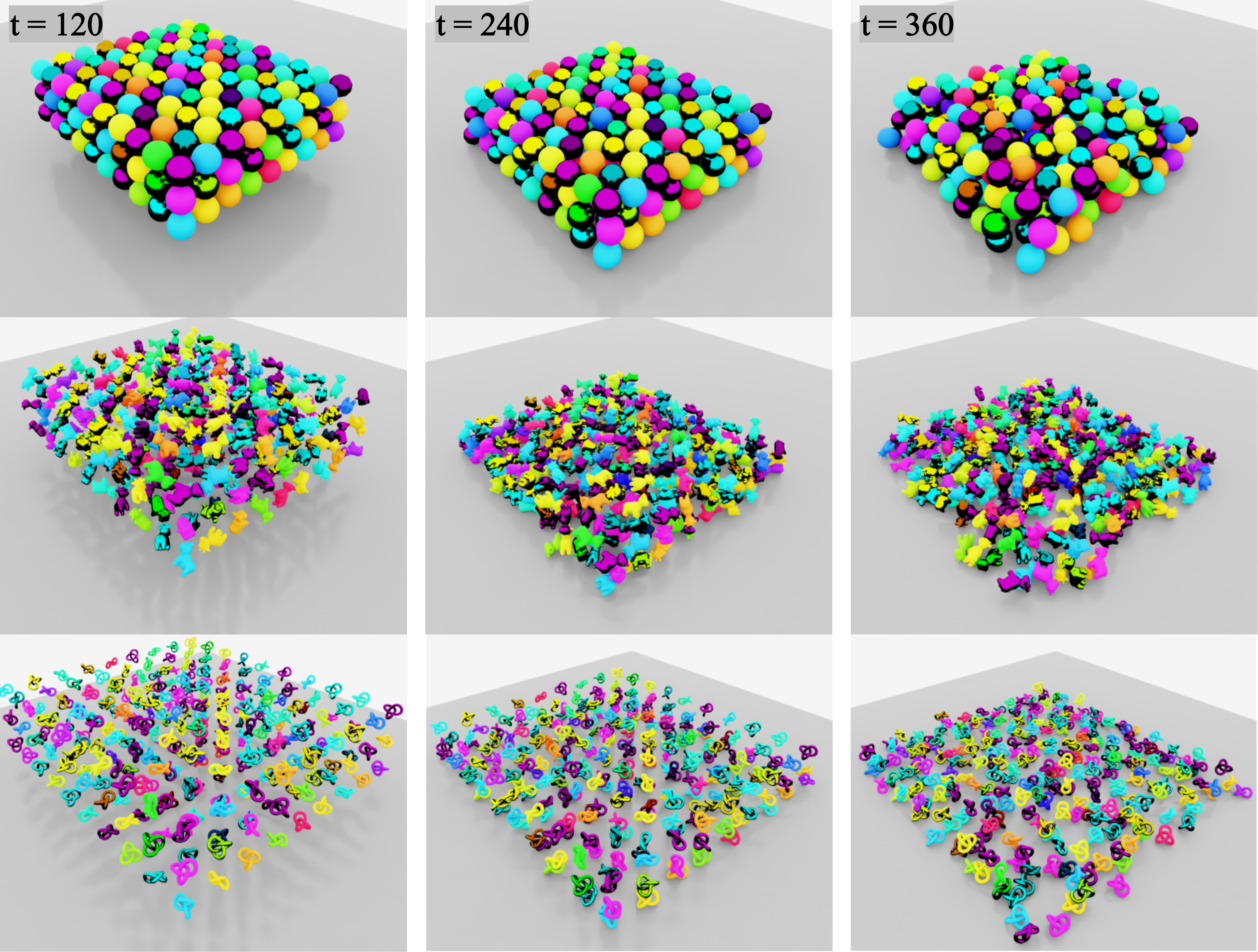}
    \caption{Qualitative results of three large-scale scenarios.}
    \label{fig:large-scale}
    \vspace{-0.2in}
\end{wrapfigure}

\noindent\textbf{Baseline implementation details.}
We reproduce RigidFormer for our experiments rather than
using runtime measurements reported in the original paper.
Although we follow its proposed architecture and inference
pipeline, implementation-level differences may affect
the measured throughput. Therefore, the reported
RigidFormer runtime should be interpreted as the
performance of our reimplementation rather than a
definitive measurement of the original implementation.

For HOPNet, we introduce additional implementation
optimizations to improve inference efficiency while
preserving its underlying modeling approach. Consequently,
the reported HOPNet throughput reflects our optimized
implementation rather than its unmodified implementation.

All three methods are benchmarked on the same GPU and
test trajectories. Nevertheless, the reported runtime
comparison should be interpreted in light of these
implementation differences.

\begin{figure}[thb]
    \centering
    \begin{minipage}[t]{0.49\linewidth}
        \centering
        \includegraphics[width=\linewidth]{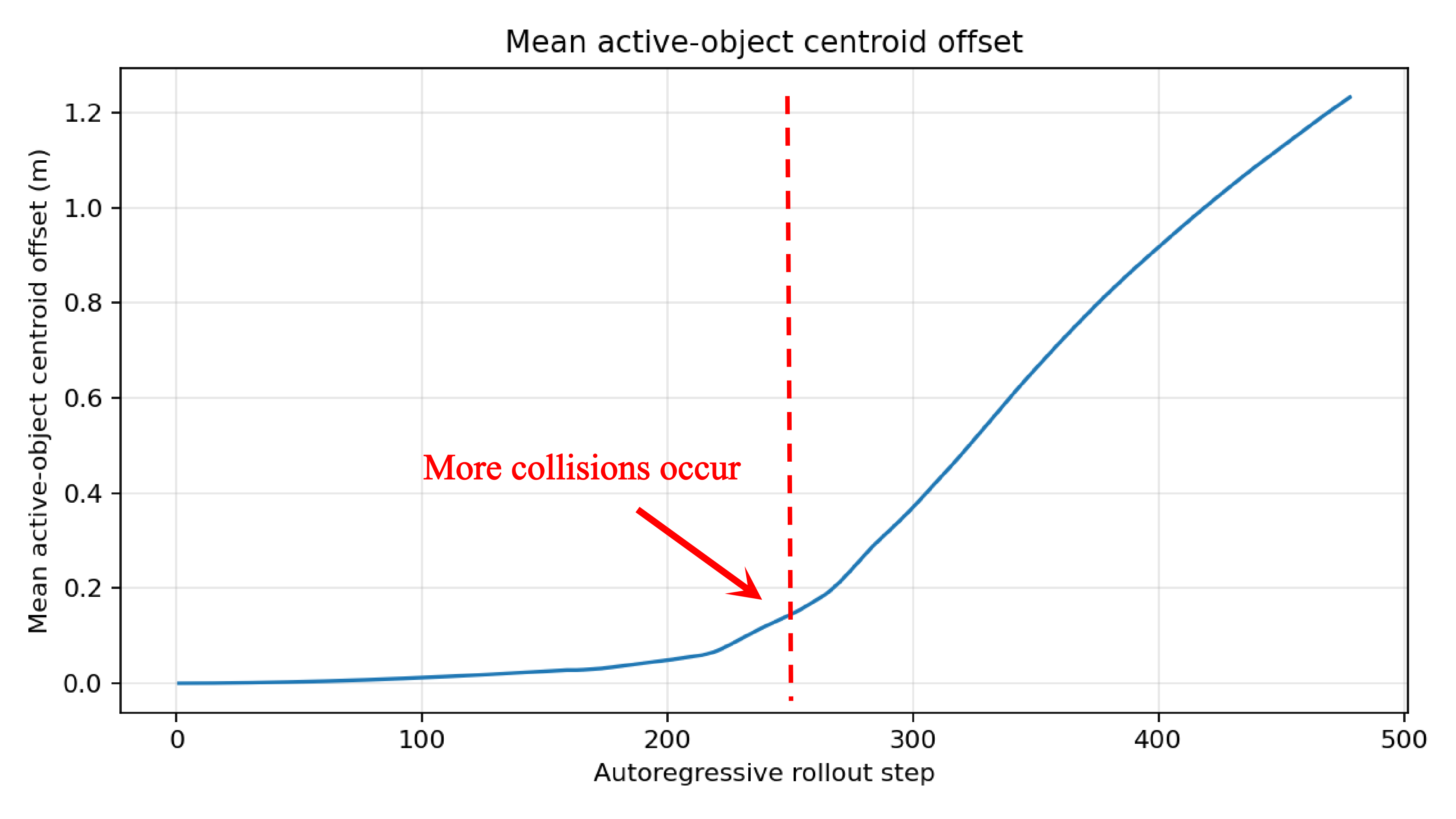}

        \small (a) Position offset in the centroid of active objects.
    \end{minipage}%
    \hfill%
    \begin{minipage}[t]{0.49\linewidth}
        \centering
        \includegraphics[width=\linewidth]{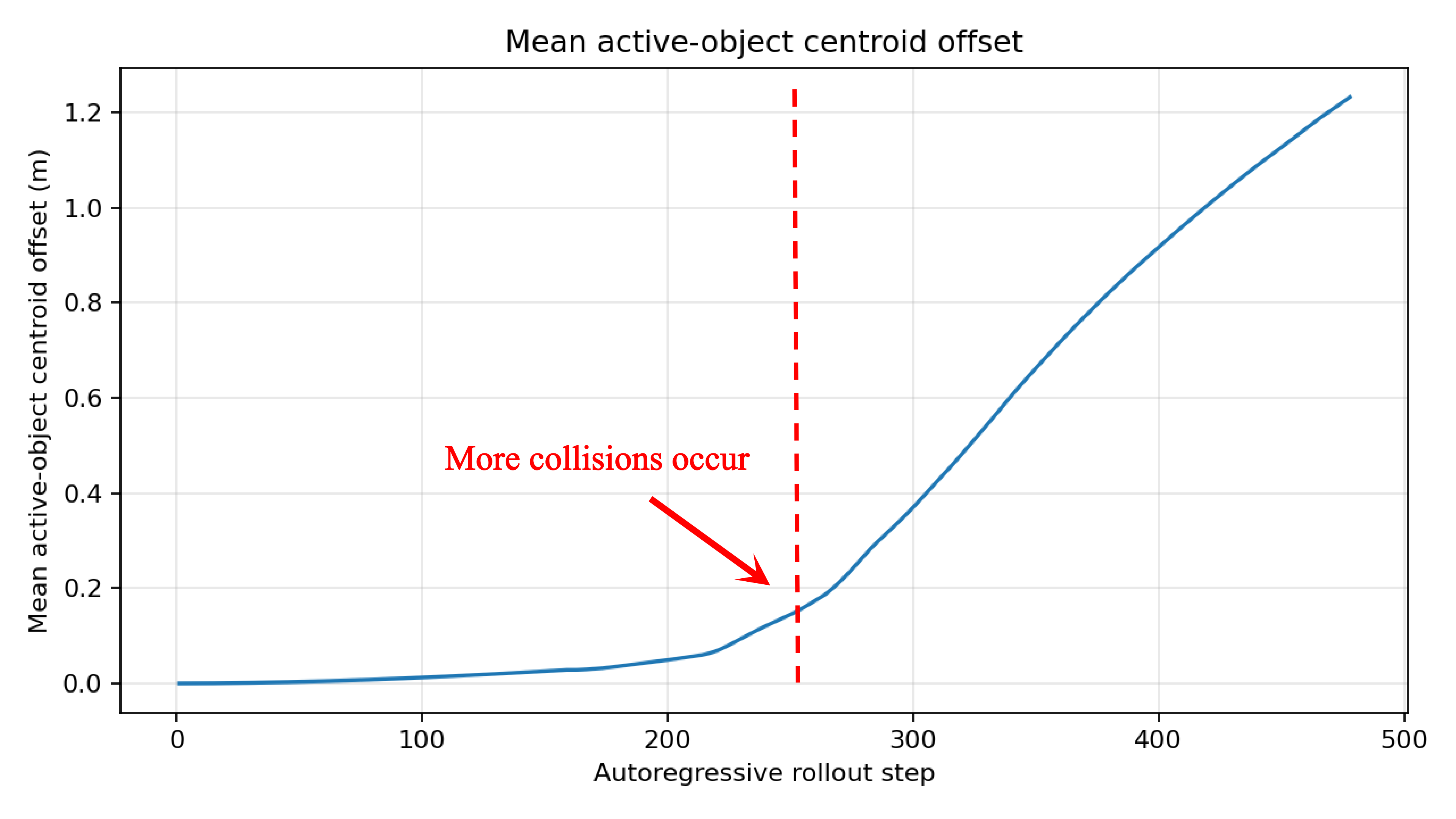}

        \small (b) Orientation Offset in the centroid of active objects.
    \end{minipage}

    \caption{
        \textbf{Temporal error accumulation in large-scale scenes.}
        Position and orientation errors throughout autoregressive
        rollouts of three 270-object scenarios. The dashed line
        indicates an approximate transition point after which
        object collisions become more frequent.
    }
    \label{fig:scalability_temporal}
\end{figure}

\subsection{Temporal Error Accumulation in Large-Scale Scenes}
\label{app:scalability_temporal}

To further analyze long-horizon behavior in the
270-object scenarios (Fig.~\ref{fig:large-scale}), we record the temporal evolution
of prediction errors throughout autoregressive rollouts.

Figure~\ref{fig:scalability_temporal} shows the position
and orientation errors as functions of rollout step.
In each plot, the dashed vertical line marks an approximate
transition point after which collisions become more frequent, based on qualitative observation of the scene dynamics.

As shown in the figure, prediction errors remain relatively
small during the earlier stage of the rollout, and then
increase more rapidly after the onset of more frequent
multi-object interactions. This trend is consistent across
the large-scale scenarios and helps explain why we report
errors not only at 120 frames, but also at longer horizons
of 240 and 360 frames in Table~\ref{tab:scalability}.

These results complement the endpoint evaluations in the
main paper by illustrating how prediction errors accumulate
over time as long-horizon autoregressive simulation becomes
increasingly challenging.

\begin{figure}[t]
    \centering
    \includegraphics[width=\linewidth]{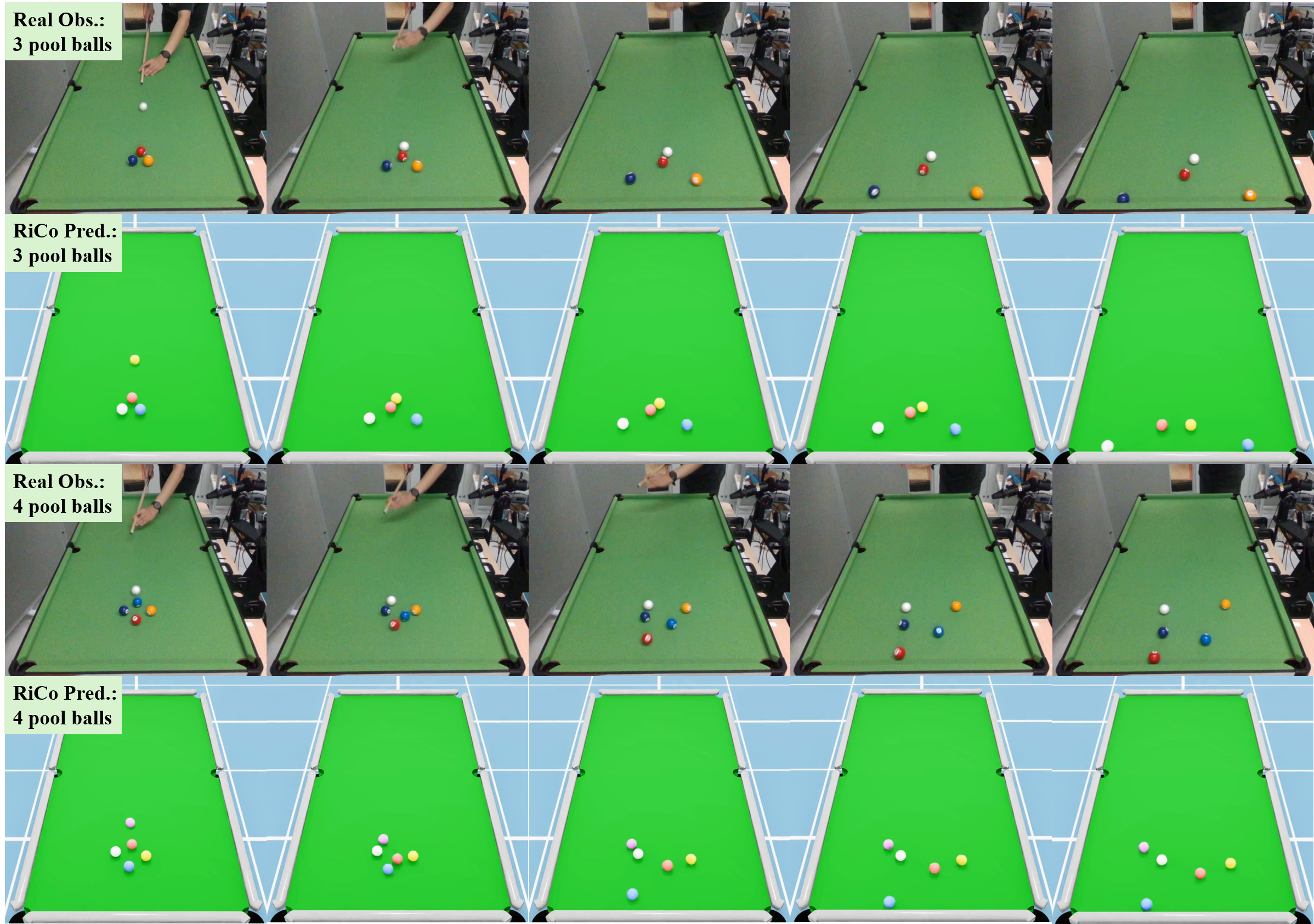}
    \caption{Illustrating 10 pairs qualitative comparison of real-world observations (upper, labeled by Real Obs.) and autoregressive predictions by \methodname\ (lower, labeled by RiCo Pred.) in three- and four-ball collision scenarios. \methodname's predictions are initialized by the balls’ observed positions and estimated velocities in the real-world experiments, then replayed in the simulated environment. Under this setting, smaller distances between corresponding predicted and observed positions indicate greater prediction accuracy.}
    \label{fig:real_world}
\end{figure}

\subsection{Real-world Experimental Setup and Simulated Training}
\label{app:sim2real}

\textbf{Real-world experimental setup.}
We conduct real-world experiments on a standard billiard table
with an effective playing area of $2.54 \times 1.27$\,m.
Each billiard ball has a radius of $33.5$\,mm.
We consider multi-ball collision scenarios involving
three and four balls, capturing their motion and
interactions through camera-based observations.

\textbf{Simulated training environment.}
To investigate zero-shot sim-to-real transfer,
we construct a corresponding simulated billiard
environment with geometric configurations comparable
to those of the real-world setup.
The simulation reproduces the table dimensions
and ball geometry used in the experiments.

To generate diverse collision dynamics, we randomize
the initial conditions of the simulated training
scenarios. Specifically, the initial position of each
ball is randomly perturbed within $\pm 0.25$\,m,
while the initial speed of the moving ball is
sampled from $1.5$--$2.5$\,m/s.
Its initial moving direction is additionally
perturbed within $\pm 5^\circ$ relative to the
nominal direction.
These randomized configurations expose the model
to different contact locations, collision angles,
and subsequent multi-object interactions.

\methodname\ is trained exclusively on 960 simulated trajectories with a duration 2s, without incorporating any real-world observations during training or performing real-world fine-tuning.

\textbf{Zero-shot evaluation protocol.}
We evaluate the trained model on five independent
real-world experimental sessions, comprising
23 valid ball samples in total.
Real-world ball configurations are reconstructed
using camera-estimated positions.
Starting from the observed initial states,
the model predicts subsequent object motion
through autoregressive rollout.

We compare the predicted ball trajectories
against the corresponding real-world observations
and report position RMSE at evaluation frames
50, 75, and 100.
For reference, we also evaluate trajectories
generated by the physics engine.
Both sets of predictions are evaluated against
the real-world observations.

\end{document}

%% file: math_commands.tex
\usepackage{amsmath,amsfonts,bm}

\def\eqref#1{equation~\ref{#1}}
\def\1{\bm{1}}

\DeclareMathAlphabet{\mathsfit}{\encodingdefault}{\sfdefault}{m}{sl}
\SetMathAlphabet{\mathsfit}{bold}{\encodingdefault}{\sfdefault}{bx}{n}